\documentclass{article}
\usepackage[utf8]{inputenc}

\usepackage[preprint]{neurips_2026}

\usepackage[utf8]{inputenc} 
\usepackage[T1]{fontenc}    
\usepackage{hyperref}       
\usepackage{url}            
\usepackage{booktabs}       
\usepackage{amsfonts}       
\usepackage{nicefrac}       
\usepackage{microtype}      
\usepackage{xcolor}         
\usepackage{multirow}

\usepackage{microtype}
\usepackage{graphicx}
\usepackage{amsmath} 
\usepackage{enumitem}
\usepackage{titlesec}
\titlespacing*{\section}{0pt}{8pt}{4pt}
\titlespacing*{\subsection}{0pt}{6pt}{3pt}
\titlespacing*{\subsubsection}{0pt}{4pt}{2pt}

\title{UniT:  Toward a Unified Physical Language for Human-to-Humanoid Policy Learning \\ and World Modeling}

\author{
  Boyu Chen$^{1,2,*}$ \quad 
  Yi Chen$^{1,3,*}$ \quad
  Lu Qiu$^{3}$ \quad
  Jerry Bai$^{1}$ \quad
  Yuying Ge$^{1,\dagger}$ \quad 
  Yixiao Ge$^1$ \\
  $^1$XPENG Robotics \quad
  $^2$Tsinghua University \quad
  $^3$The University of Hong Kong \\
  \small{\color{blue} \url{https://xpeng-robotics.github.io/unit/}}
}

\begin{document}

\maketitle

\let\thefootnote\relax\footnotetext{$^*$ Equal contribution.}
\let\thefootnote\relax\footnotetext{$\dagger$ Corresponding author.}

\begin{figure}[!htbp]
  \centering
    \includegraphics[width=1.0\textwidth]{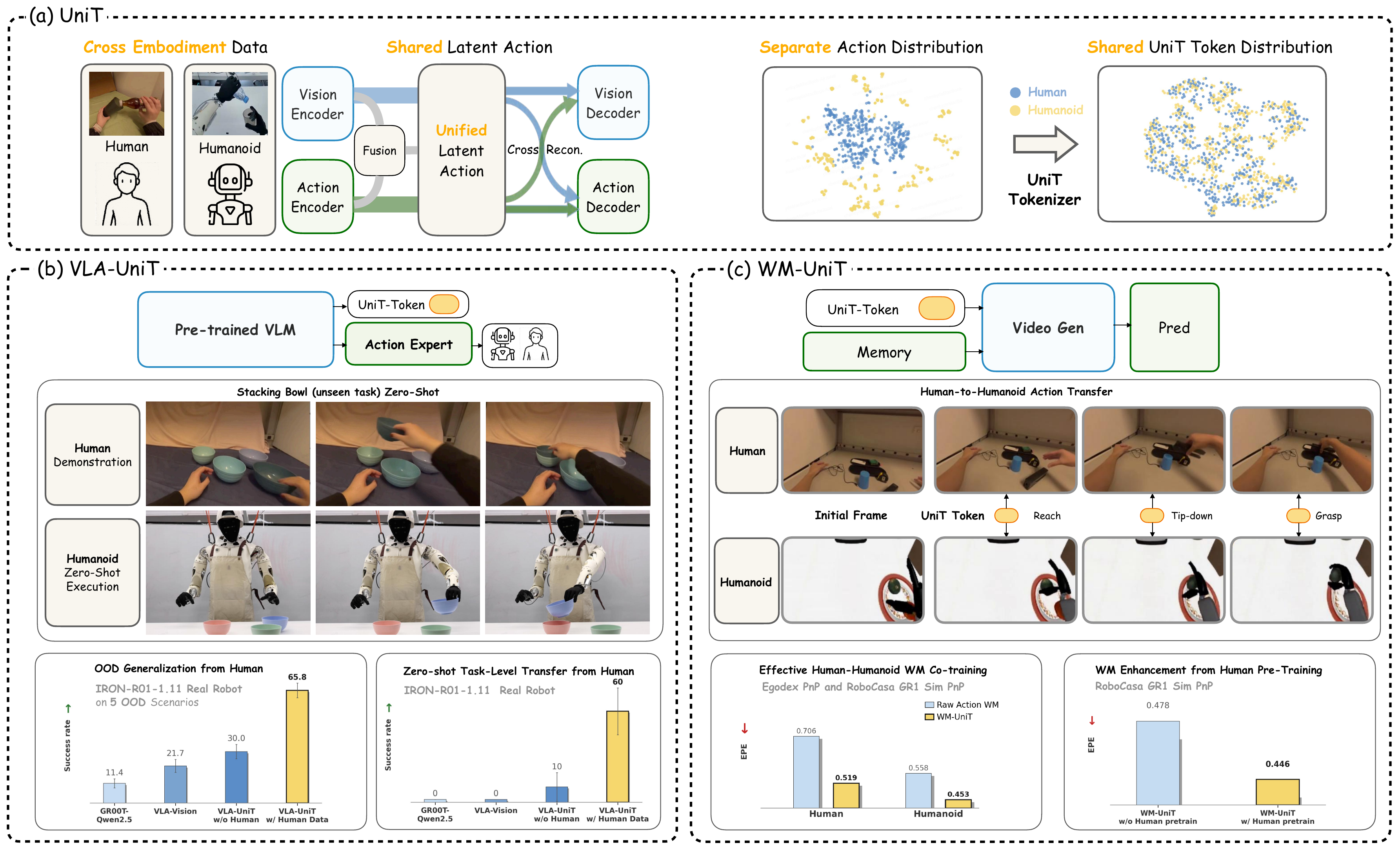}
    \caption{
    Overview of the UniT Framework, which establishes a \textbf{unified physical language} to bridge the human-humanoid chasm. 
    (a) \textbf{Unified Tokenization:} A tri-branch cross-reconstruction mechanism projects heterogeneous actions into a shared discrete latent space (verified via t-SNE) for extracting unified physical intents.
    (b) \textbf{Policy Learning}: Through predicting these unified tokens, VLA-UniT effectively leverages diverse human data to achieve robust OOD generalization and \textit{zero-shot task transfer} on humanoid robot.
    (c) \textbf{World Modeling}: By aligning cross-embodiment dynamics through unified token conditioning, WM-UniT enables direct human-to-humanoid action transfer, effectively leveraging human priors to improve controllability in humanoid video generation.
    }
    \label{fig:teaser}
\end{figure}

\begin{abstract}
Scaling humanoid foundation models is bottlenecked by the scarcity of robotic data. While massive egocentric human data offers a scalable alternative, bridging the cross-embodiment chasm remains a fundamental challenge due to kinematic mismatches. We introduce \textbf{UniT} (\textbf{Uni}fied Latent Action \textbf{T}okenizer via Visual Anchoring), a framework that establishes a unified physical language for human-to-humanoid transfer. Grounded in the philosophy that heterogeneous kinematics share universal visual consequences, UniT employs a tri-branch cross-reconstruction mechanism: actions predict vision to anchor kinematics to physical outcomes, while vision reconstructs actions to filter out irrelevant visual confounders. Concurrently, a fusion branch synergies these purified modalities into a shared discrete latent space of embodiment-agnostic physical intents. We validate UniT across two paradigms: 1) \textbf{Policy Learning (VLA-UniT):} By predicting these unified tokens, it effectively leverages diverse human data to achieve state-of-the-art data efficiency and robust out-of-distribution (OOD) generalization on both humanoid simulation benchmark and real-world deployments, notably demonstrating \textit{zero-shot task transfer}. 2) \textbf{World Modeling (WM-UniT):} By aligning cross-embodiment dynamics via unified tokens as conditions, it realizes direct human-to-humanoid action transfer. This alignment ensures that human data seamlessly translates into enhanced action controllability for humanoid video generation. Ultimately, by inducing a highly aligned cross-embodiment representation 
(empirically verified by t-SNE visualizations revealing the convergence of human and humanoid features into a shared manifold),
UniT offers a scalable path to distill vast human knowledge into general-purpose humanoid capabilities.
\end{abstract}

\section{Introduction}

Scaling foundation models for humanoids in both policy learning and world modeling is fundamentally bottlenecked by scarce high-quality robotic data. Massive, structured human motion sequences from low-cost capture provide a scalable alternative rich in physical interaction priors, but leveraging them requires bridging a major cross-embodiment gap\cite{cai2025n}. Biomechanical and hardware differences create heterogeneous state-action spaces with mismatched degrees of freedom (DoF) and control paradigms. Traditional pipelines rely on \textbf{motion retargeting}\cite{tao2025dexwild, yang2025egovla}, which uses complex kinematic solvers to map human motions to specific robots. This case-by-case process is labor-intensive, unscalable, and often physically inconsistent. We therefore need a data-driven \textbf{unified physical language} that projects heterogeneous data into a shared latent action space.

\begin{figure}[!t]
    \centering
    \includegraphics[width=0.95\textwidth]{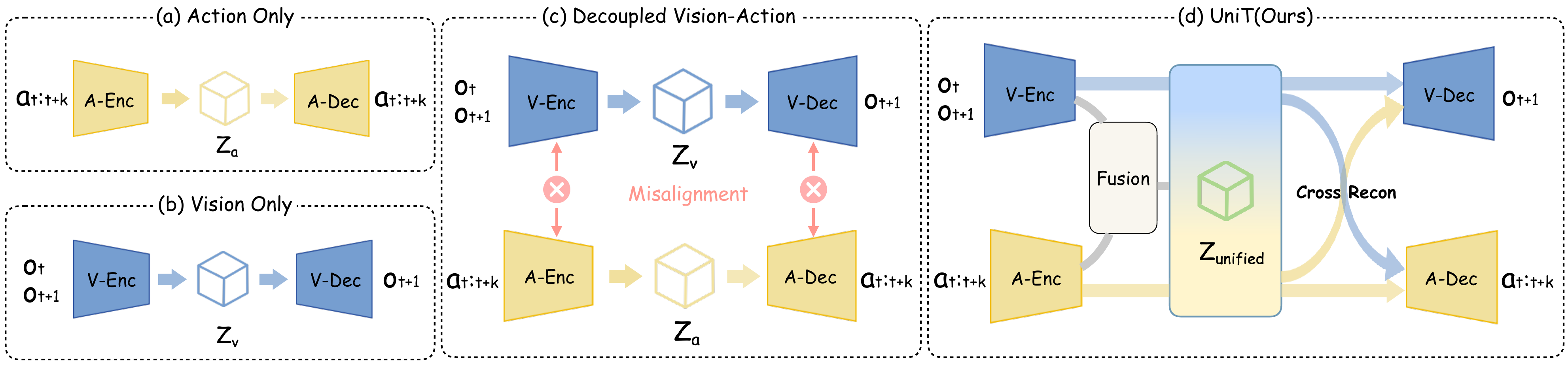}
    \caption{
    \textbf{Comparison of Latent Action Architectures.} 
    \textbf{(a) Action-Only:} Relies on action reconstruction ($Z_a$), suffering from distribution misalignment without visual grounding. 
    \textbf{(b) Vision-Only:} Infers representations ($Z_v$) from pixels, which entangles with low-level appearances and misses fine-grained physical details. 
    \textbf{(c) Decoupled Vision-Action:} Encodes modalities into disconnected spaces ($Z_v, Z_a$), lacking the explicit alignment needed to extract shared physical concepts. 
    \textbf{(d) UniT (Ours):} Fuses heterogeneous data into a shared space ($Z_{unified}$) via \textbf{cross-reconstruction}. This strict cross-modal alignment ensures tokens capture universal physical intentions, forming a robust unified physical language.
    }
    \label{fig:related_work}
\end{figure}

Although recent literature explores unified representations, critical limitations remain (Fig.~\ref{fig:related_work}). \textbf{Action-only} methods~\cite{lee2024behavior, pertsch2025fast, vuong2025action, mete2407quest, wang2025vq} rely exclusively on proprioceptive reconstruction, often suffering from severe distribution shifts between humans and robots due to the lack of external grounding. Conversely, emerging \textbf{latent action} frameworks are predominantly \textbf{vision-only}~\cite{chen2025moto, ye2024latent, bu2025univla, bu2025agibot_iros}, inferring intent directly from pixels. While bypassing kinematic mismatches, these representations are prone to entangling low-level appearance confounders (e.g., textures and lighting). This entanglement limits fine-grained physical execution and leaves the structural priors of human pose data underexploited. Furthermore, while some paradigms~\cite{fu2025metis} incorporate \textbf{both vision and action}, they typically employ independent tokenizers for each modality. This results in disjoint vocabularies without deep representational unification, failing to establish a truly universal medium for control. Empirically, many existing latent-action systems are confined to fixed single- or dual-arm setups with simple grippers, leaving their scalability to dexterous humanoid largely underexplored.

To address these challenges, we introduce \textbf{UniT} (\textbf{Uni}fied Latent Action \textbf{T}okenizer via Visual Anchoring). Our design is driven by a critical insight into cross-embodiment alignment: while human and humanoid kinematics differ in structural DoFs and contain embodiment-specific noise, the physical outcomes of their intents share a consistent visual representation. Therefore, visual observations can serve as a universal anchor to ground and align disparate kinematic spaces. Building on this principle, UniT functions as a cross-modal information bottleneck to distill the underlying physical intent. The tokenizer concurrently extracts three coupled representations: a \textit{temporal-visual} feature from consecutive frames, a \textit{kinematic} feature from corresponding inter-frame actions, and a \textit{fused visuo-motor} feature. Instead of treating these as isolated streams, we enforce a rigorous \textbf{cross-reconstruction objective}, compelling each representation to independently decode both the visual transitions and the low-level actions.

This mechanism operationalizes the concept of \textbf{visual anchoring}. By forcing kinematic features to reconstruct visual transitions, heterogeneous actions are anchored to their actual physical consequences in the environment. This constraint prevents the network from merely memorizing embodiment-specific kinematics and filters out visually unobservable artifacts. Conversely, compelling visual features to reconstruct kinematics strips away low-level appearance confounders, such as lighting and irrelevant backgrounds, that do not contribute to physical motion. Uncorrelated noise from either domain is discarded during optimization, preserving only the essential intersection of both modalities: the embodiment-agnostic physical intent. As a result, UniT generates deeply integrated visuo-motor tokens that synergize visual and kinematic information into a unified latent action space. This strict alignment also ensures that the independent visual and kinematic branches extract structured cross-modal tokens, maintaining robust representations even in the absence of a modality during deployment. Ultimately, these visually-anchored discrete tokens serve as a \textbf{universal physical language}, providing a stable foundation for transferring intent across different robot morphologies.

To evaluate this unified language, we deploy UniT in two embodied-AI paradigms:

First in \textbf{Policy Learning.} We introduce \textbf{VLA-UniT} by integrating UniT into Vision-Language-Action architectures. Instead of fitting raw actions across large distribution gaps, VLA-UniT predicts UniT tokens in the shared latent space, and a lightweight flow head then generates embodiment-specific actions for execution. We evaluate on the RoboCasa GR1 benchmark and a real humanoid. VLA-UniT outperforms state-of-the-art baselines and improves data efficiency and out-of-distribution (OOD) generalization using diverse human data, and shows zero-shot task transfer with emergent upper-body coordination (e.g., waist rotation) on unseen tasks.

Second in \textbf{World Modeling.} We propose \textbf{WM-UniT}, which uses UniT tokens as universal conditions instead of raw actions. Because these tokens absorb physical priors during joint dynamics training, they improve prediction consistency. Rollout validations show that pre-training on large-scale human data aligns fine-grained human and humanoid actions, enabling transfer of physical dynamics across embodiments and improving downstream humanoid control generation.

Beyond downstream performance, UniT offers structural benefits. It forms a shared cross-embodiment latent space and also encourages downstream models to align their internal features. In both VLA policies and world models, UniT tokens induce more aligned cross-embodiment context representations. The tokenizer also provides useful denoising: encoding and decoding noisy captured actions filters perturbations and recovers cleaner, more executable trajectories.

Our main contributions are summarized as follows:
\begin{itemize}[nosep,leftmargin=*]
    \item \textbf{The Unified Tokenizer (UniT):} We propose a visual-anchored tri-branch tokenizer with cross-reconstruction that maps heterogeneous actions into a shared discrete latent space, yielding robust cross-embodiment alignment and action denoising.
    \item \textbf{Generalizable Policy Execution (VLA-UniT):} We integrate UniT into VLA architectures and achieve strong data efficiency, superior OOD generalization, and zero-shot task transfer for humanoids in simulation and the real world.
    \item \textbf{Unified World Modeling (WM-UniT):} We use UniT tokens as universal conditions for world models, showing that co-training on human and humanoid data improves dynamics prediction and downstream control generation.
\end{itemize}

\section{Related Work}

\textbf{Learning from Human Data.} Leveraging human data for robot learning has attracted growing attention as a scalable alternative to costly robot demonstrations. A prominent line of work pre-trains visual representations from egocentric human videos~\cite{nair2022r3m, ma2022vip, radosavovic2023real, wang2024scaling}, showing positive transfer to downstream manipulation. However, most of these approaches focus on unsupervised visual learning without exploiting fine-grained hand or wrist pose information, limiting their utility for dexterous upper-body manipulation. More recent methods co-train unified policies on human and robot demonstrations through explicit alignment or motion information~\cite{qiu2025humanoid, yuan2025motiontrans, kareer2025egomimic, tao2025dexwild, bi2026h}. However, as illustrated in~\cite{cai2025n, kareer2025emergence}, co-training on mixed embodiment data end-to-end forces the model to fit fundamentally different action distributions simultaneously, often leading to embodiment-specific shortcuts rather than shared representations. To mitigate this, another line of work introduces motion retargeting. EgoVLA~\cite{yang2025egovla} and In-n-On~\cite{cai2025n} predict unified human wrist and hand actions and then apply inverse kinematics to map them onto robot joint configurations. However, retargeted actions often misalign with the original visual observations, and the IK-based pipeline remains case-by-case and difficult to scale across diverse morphologies. UniT addresses these challenges by learning a shared latent space that absorbs embodiment differences at the representation level, bypassing the need for explicit retargeting or action-space unification.

\textbf{Latent Action Representations.} A growing body of work explores latent action representations for robot learning, which we categorize by the modality they encode.
\textbf{Action-only} methods~\cite{lee2024behavior, pertsch2025fast, vuong2025action, mete2407quest, wang2025vq} learn discrete or continuous autoencoders over raw action trajectories to produce compact representations for policy learning. Among them, VQ-BeT~\cite{lee2024behavior} and FAST~\cite{pertsch2025fast} demonstrate that structured tokenization improves behavior generation. However, without external grounding, these representations reflect embodiment-specific kinematics and struggle to align heterogeneous action distributions, limiting cross-embodiment transferability.
\textbf{Vision-only} methods~\cite{chen2025moto, ye2024latent, bu2025univla, bu2025agibot_iros} infer latent actions directly from visual observations to bypass kinematic mismatches. Moto~\cite{chen2025moto} and LAPA~\cite{ye2024latent} learn latent motion tokens from video, while UniVLA~\cite{bu2025univla} uses vision-derived latent actions for cross-embodiment policy learning. While this offers cross-domain potential, such representations tend to entangle low-level appearance factors and miss fine-grained motor detail, underexploiting the structural priors available in human pose data.
Villa-X~\cite{chen2025villa} partially addresses this by incorporating action reconstruction as an auxiliary target, but the unidirectional vision-to-action objective still limits the precision of the learned motor representation.
Concurrent works such as METIS~\cite{fu2025metis} and XR-1~\cite{fan2025xr} take \textbf{both vision and action as encoder inputs} but have not achieved explicit vision-action alignment. XR-1 applies KL regularization to encourage distributional proximity, which may not fully capture the fine-grained cross-modal correspondence that cross-reconstruction provides.

\textbf{Vision-Language-Action Model and Action-Conditioned World Models.} \textbf{Policy learning} and \textbf{world modeling} represent two core paradigms for embodied AI. For policy learning, VLA models~\cite{chi2023diffusion, gr00tn1_2025, pertsch2025fast, kim2025fine} integrate vision-language backbones with action generation for closed-loop control. Cross-embodiment generalist policies such as GR00T~\cite{gr00tn1_2025}, $\pi_0$~\cite{black2024pi_0}, RT-X~\cite{o2024open}, and Octo~\cite{team2024octo} train across multiple embodiments, generating raw actions via diffusion heads or predicting action tokens from proprioceptive data. For world modeling, action-conditioned video generation has emerged as a promising approach for simulating robotic dynamics. IRASim~\cite{zhu2025irasim}, Ctrl-World~\cite{guo2025ctrl}, and WPE~\cite{quevedo2025evaluating} explore controllable generation conditioned on robot actions, building upon video foundation models such as Cosmos~\cite{nvidia2025worldsimulationvideofoundation}. Across both paradigms, the heterogeneity of action spaces across embodiments remains a key bottleneck, as most existing systems are confined to single-embodiment or single-arm gripper settings with limited cross-embodiment validation. UniT provides a unified token interface for both paradigms, projecting heterogeneous actions into a shared latent space that serves as a prediction target for VLA and a conditioning signal for world models, enabling scalable human-to-humanoid transfer across both policy learning and world modeling.

\section{Methodology}

\begin{figure}[!t]
    \centering
    \includegraphics[width=0.95\textwidth]{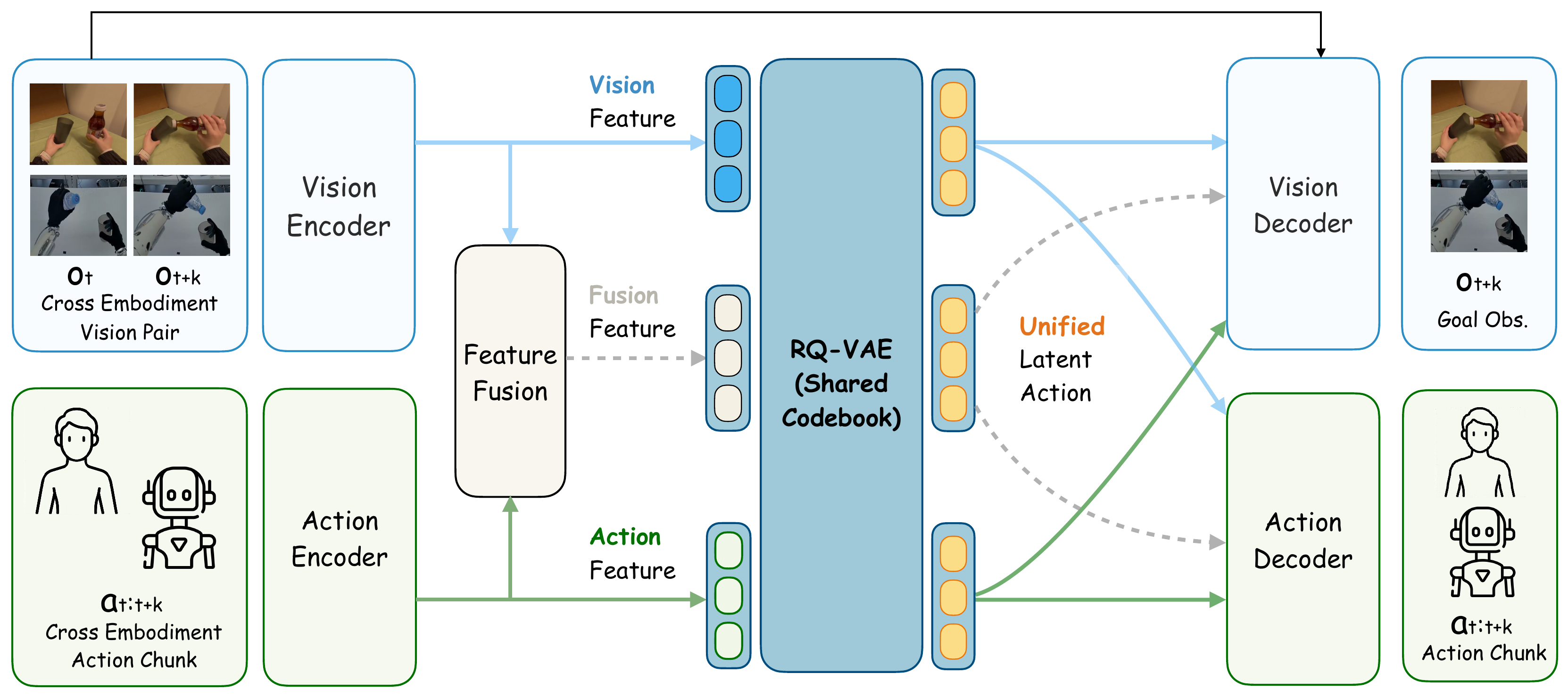}
    \caption{
    \textbf{Architecture of UniT.} Heterogeneous cross-embodiment vision pairs $(o_t, o_{t+k})$ and action chunks $a_{t:t+k}$ are encoded into vision, action, and fused features via tri-branch encoders. A \textbf{shared RQ-VAE codebook} quantizes all three branches into a unified discrete space, yielding embodiment-agnostic \textbf{Unified Latent Action} tokens. Both vision and action decoders reconstruct from these shared tokens, enforcing \textbf{cross-reconstruction} alignment.
    }
    \label{fig:tokenizer}
\end{figure}

\subsection{Overview}
\label{sec:overview}

Our goal is to establish a unified physical language that bridges heterogeneous human and humanoid action spaces for cross-embodiment transfer. We consider human and humanoid demonstrations as sequences of observations, embodiment-specific states, and actions, $(o_t, s_t, a_t)$. We first present \textbf{UniT} (Sec.~\ref{sec:unit}), a visual-anchored latent action tokenizer built upon a tri-branch architecture. By jointly modeling visual transitions, actions, and fused visuo-motor features, UniT produces discrete tokens that capture embodiment-agnostic physical intent and serve as a unified interface across embodiments.

Leveraging this shared token representation, we deploy UniT in two complementary embodied-AI paradigms. \textbf{VLA-UniT} (Sec.~\ref{sec:vla-unit}) models the policy decision process by predicting future action chunks from the current observation and state through UniT token prediction and embodiment-specific action generation. \textbf{WM-UniT} (Sec.~\ref{wm-unit}) models the dynamics prediction process by predicting future visual observations from the current observation and action conditions, using UniT features as a universal control interface instead of embodiment-specific raw actions.

\subsection{UniT: Unified Latent Action Tokenizer via Vision Anchoring}
\label{sec:unit}

Given an observation transition $(o_t, o_{t+k})$, current state $s_t$, and action chunk $a_{t:t+k}$, UniT learns a discrete latent action representation that maps heterogeneous human and humanoid behaviors into a shared token space. The resulting tokens serve as a unified embodiment-agnostic interface for downstream policy learning and world modeling. To achieve this, UniT is built upon a tri-branch architecture that jointly models visual transitions, actions, and fused visuo-motor features under cross-reconstruction (Fig.~\ref{fig:tokenizer}).

\paragraph{Tri-Branch Encoding.}
UniT uses three parallel branches. Each branch employs a transformer encoder with learnable queries to summarize modality-specific inputs into a compact latent representation before quantization.
\begin{itemize}[nosep,leftmargin=*]
  \item \textbf{Visual branch} $E_v$: operates as an inverse dynamics model (IDM), taking frozen DINOv2~\cite{oquab2023dinov2} features of the observation pair $(o_t, o_{t+k})$ as input, following the visual feature design adopted in UniVLA~\cite{bu2025univla}, and producing a latent representation of the underlying physical transition. The domain-invariant nature of DINOv2 provides a stable visual anchor across embodiments.
  \item \textbf{Action branch} $E_a$: encodes current state $s_t$ and action chunk $a_{t:t+k}$. Because human and humanoid embodiments differ in control mode, action parameterization, and degrees of freedom, raw actions are first padded to a unified maximum length and projected by embodiment-specific MLPs, then summarized into a compact latent control representation.
  \item \textbf{Fusion branch} $E_m$: takes the branch features from vision and action as input and produces a fused visuo-motor latent representation, capturing complementary cross-modal structure for more compact and robust tokens.
\end{itemize}

\paragraph{Shared Discrete Quantization.}
Continuous latents from all three branches are quantized via Residual Quantization (RQ-VAE) with a shared codebook $\mathcal{C}$:
\begin{equation}
    \hat{z}_i = \operatorname{RQ}(z_i;\,\mathcal{C}), \quad i \in \{v, a, m\}.
\end{equation}
The shared codebook ensures that tokens from all branches and all embodiments reside in a unified discrete space. Residual quantization progressively refines the approximation through multiple codebook levels, capturing both coarse physical intent and fine-grained motion detail within a compact representation.

\paragraph{Cross-Reconstruction.}
The core mechanism of UniT is cross-reconstruction: every quantized token $\hat{z}_i$ is decoded by both a shared visual decoder $D_v$ and an embodiment-specific action decoder $D_a$:
\begin{equation}
    \hat{f}_{t+k}^{(i)} = D_v(\hat{z}_i,\, f_t), \qquad \hat{a}_{t:t+k}^{(i)} = D_a(\hat{z}_i,\, s_t),
\end{equation}
where $f_t$ denotes the DINOv2 feature of $o_t$. The visual decoder operates as a forward dynamics model (FDM) conditioned on the current observation, supervised by cosine similarity against $f_{t+k}$; the action decoder is conditioned on the current state $s_t$ to reconstruct the action chunk. This design encourages the token to capture relative physical change, rather than memorizing absolute configurations, which benefits cross-embodiment transfer because different embodiments can still share analogous transition patterns despite vastly different state spaces.

This bidirectional constraint realizes the principle of visual anchoring: \textit{vision provides a universal physical anchor across embodiments}. While heterogeneous action spaces are inherently incomparable, visually similar task outcomes can still be shared across embodiments. Through a shared visual decoder, tokens producing similar physical effects can converge to nearby codebook entries regardless of embodiment, anchoring heterogeneous motor representations into a unified manifold.

\paragraph{Training Objective.}
The total loss aggregates cross-reconstruction and quantization terms across all three branches:
\begin{equation}
    \mathcal{L} = \sum_{i \in \{v,a,m\}} \Big[\lambda_v\, \mathcal{L}_{\text{cos}}(\hat{f}_{t+k}^{(i)},\, f_{t+k}) + \lambda_a\, \mathcal{L}_{\text{act}}(\hat{a}_{t:t+k}^{(i)},\, a_{t:t+k})\Big] + \mathcal{L}_{\text{RQ}},
\end{equation}
where $\mathcal{L}_{\text{cos}}$ is the cosine similarity loss for visual feature reconstruction, $\mathcal{L}_{\text{act}}$ is the action reconstruction loss, and $\mathcal{L}_{\text{RQ}}$ is the RQ-VAE commitment loss. For downstream deployment, VLA-UniT leverages fusion-branch tokens that integrate visual and motor understanding for policy prediction (Sec.~\ref{sec:vla-unit}), while WM-UniT uses action-branch features as the control interface, since future frames must be generated by the world model itself and are therefore unavailable at the input side (Sec.~\ref{wm-unit}).

\begin{figure}[!t]
    \centering
    \includegraphics[width=1.0\textwidth]{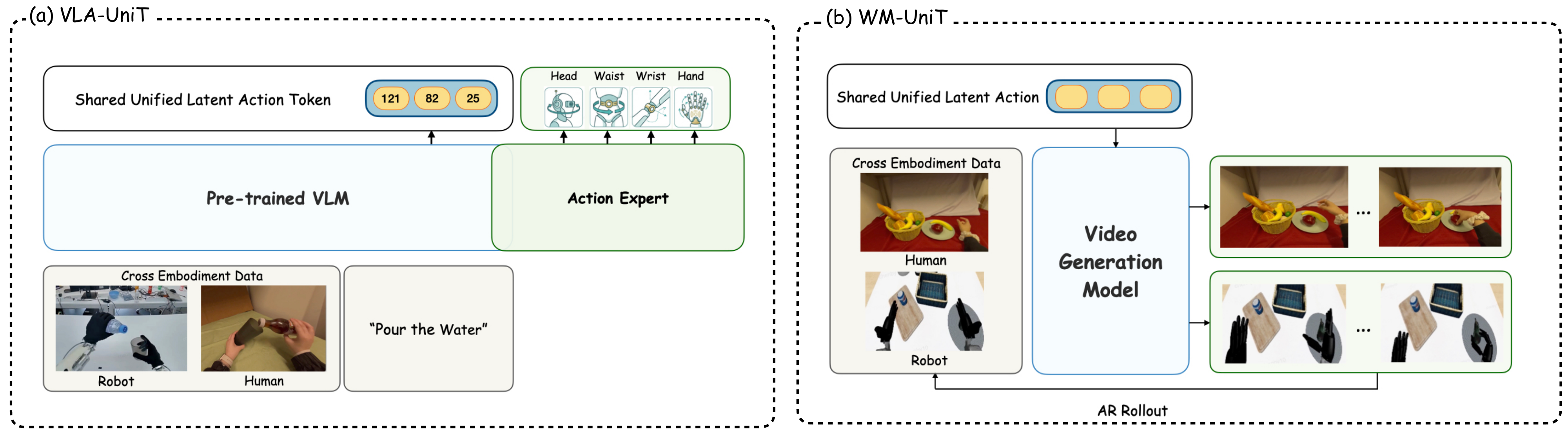}
    \caption{
    \textbf{Downstream Applications of UniT Tokens.}
    \textbf{(a) VLA-UniT:} Integrates UniT token prediction into a VLA architecture. Rather than direct action regression, the model predicts Unified Latent Action tokens in the shared space, while a lightweight action expert generates embodiment-specific controls from the same vision-language context, covering head pose, waist pose, wrist pose, and hand pose.
    \textbf{(b) WM-UniT:} Employs UniT tokens as universal action conditions for world modeling instead of embodiment-specific raw actions, generating future frames with autoregressive rollout.
    }
    \label{fig:vla_wm}
\end{figure}

\subsection{VLA-UniT: Cross Embodiment Policy Learning via UniT}
\label{sec:vla-unit}

Given the current observation $o_t$, state $s_t$, and language instruction $\ell$, VLA-UniT models the policy decision process by predicting a future action chunk $a_{t:t+H}$. Built upon the GR00T n1.5 framework~\cite{gr00tn1_2025} with Qwen2.5-VL~\cite{Qwen2.5-VL} as the vision-language backbone, it uses UniT tokens as a structured cross-embodiment prediction target for the VLM, while the action expert generates embodiment-specific controls from the same vision-language context. We therefore decompose policy learning into UniT token prediction and flow-matching action generation.

\paragraph{UniT Token Prediction.} Learnable queries $q_t$ are appended to the VLM sequence. Conditioned on the current observation $o_t$ and language instruction $\ell$, the VLM predicts UniT discrete codes through these queries. The target codes are obtained by encoding the ground-truth observation pair, state and action chunk through the pre-trained UniT tokenizer, $c_t = \mathrm{UniT}(o_t, o_{t+k}, a_{t:t+k}, s_{t})$. Here, UniT serves as the prediction target, while the VLM vision-language context provides the conditioning signal. Since UniT tokens are discrete by construction, they naturally match the token prediction objective used for VLM training. Formally, letting $\hat{p}_t$ denote the predicted logits over UniT code indices,
\begin{equation}
    \hat{p}_t = f_{\text{VLM}}(o_t,\, \ell,\, q_t), \qquad
    \mathcal{L}_{\text{token}} = \mathrm{CE}(\hat{p}_t,\, c_t).
\end{equation}

\paragraph{Flow Matching Action Generation.} Continuous embodiment-specific actions are generated via a lightweight flow matching head. Given action chunk $A_t = [a_t, \dots, a_{t+H-1}]$, time variable $\tau \sim \mathcal{U}[0,1]$, and noise $\epsilon \sim \mathcal{N}(\mathbf{0}, \mathbf{I})$, we construct the interpolated path $A_t^\tau = \tau A_t + (1-\tau)\epsilon$. The flow head learns a velocity field $V_\theta$ conditioned on vision-language features $x_t$ from the last layer of the VLM and the current observation encoding from the VLM visual encoder, $\mathrm{Enc}(o_t)$:
\begin{equation}
    \mathcal{L}_{\text{fm}} = \mathbb{E}_{\tau, \epsilon} \left[ \left\| V_\theta(A_t^\tau \mid x_t,\, \mathrm{Enc}(o_t),\, \tau) - (A_t - \epsilon) \right\|^2_2 \right].
\end{equation}
The total objective combines both terms: $\mathcal{L}_{\text{VLA}} = \mathcal{L}_{\text{token}} + \lambda_{\text{fm}}\, \mathcal{L}_{\text{fm}}$. Importantly, both UniT token prediction and action generation are conditioned on the same VLM vision-language features $x_t$. Since UniT maps heterogeneous human and humanoid behaviors into a shared physical intent vocabulary, token prediction pulls these context features into a more unified cross-embodiment space. The action expert then generates embodiment-specific actions from this unified context, enabling cross-embodiment transfer within a single policy (Sec.~\ref{sec:experiments}).

\subsection{WM-UniT: Cross Embodiment World Modeling via UniT}
\label{wm-unit}


Given the current observation $o_t$ and action condition derived from $(s_t, a_{t:t+k})$, WM-UniT models the future prediction process by generating future visual observations. Built upon the Cosmos Predict 2.5~\cite{nvidia2025worldsimulationvideofoundation} action-conditioned video generation framework (Fig.~\ref{fig:vla_wm}b), it uses action-branch UniT features as a unified conditioning interface that replaces embodiment-specific raw actions. We therefore use UniT features as the control signal for action-conditioned video generation. Specifically, given state $s_t$ and action chunk $a_{t:t+k}$, the UniT action-branch encoder $E_a$ produces continuous pre-quantization features $\tilde{z}^{\,a}_t = E_a(s_t, a_{t:t+k})$, which are projected through an MLP and injected via cross-attention alongside the current observation $o_t$. The generation model is trained with flow matching (analogous to Sec.~\ref{sec:vla-unit}), with the velocity field applied to latent future frames $X$:
\begin{equation}
    \mathcal{L}_{\text{WM}} = \mathbb{E}_{\tau, \epsilon} \left[ \left\| V_\phi\!\left(X_t^\tau \mid o_t,\, \operatorname{MLP}(\tilde{z}^{\,a}_t),\, \tau\right) - (X_t - \epsilon) \right\|^2_2 \right].
\end{equation}
Here, $X_t$ denotes the latent representation of future frames. For long-horizon evaluation, WM-UniT supports autoregressive rollout by feeding generated frames back as observations for subsequent steps. 

UniT uses vision as the anchor to map human and humanoid actions into a shared intent space, so the action features $\tilde{z}^{\,a}_t$ provide an embodiment-agnostic control interface while naturally carrying visual-dynamics priors learned during tokenization. The world model can therefore use UniT features as a unified conditioning signal across embodiments, enabling transfer in visual generation from human data to humanoid prediction. Because the action branch itself takes only state and action as input, this conditioning does not leak future observations at deployment time.



\section{Experimental Setups}
\subsection{Benchmarks and Datasets}
\subsubsection{RoboCasa GR1 Tabletop Simulation}

We evaluate on the RoboCasa benchmark with the GR1 humanoid robot~\cite{nvidia2025gr00tn1openfoundation}. The evaluation suite comprises 24 tabletop tasks: 18 pick-and-place rearrangement tasks where the robot follows language instructions to move objects between containers, and 6 articulated tasks that involve more complex interactions such as placing objects inside and subsequently closing cabinets, drawers, or microwaves. Each task is assessed over 50 episodes in simulation.

\paragraph{Data Configurations.} We evaluate under two data regimes. \textbf{Full Data} uses 24,000 robot trajectories (1,000 per task). \textbf{Few-Shot} uses a 10\% subset of 2,400 trajectories (100 per task).

\paragraph{Learning from Human Data.} 
To examine human-to-humanoid transfer for both policy learning and world modeling, we incorporate the \texttt{basic\_pick\_place} subset of the EgoDex dataset~\cite{hoque2025egodex}, containing 27,419 trajectories of pick-and-place interactions. The human data is combined with the few-shot robot set for co-training, followed by fine-tuning exclusively on robot data.

\paragraph{Generalization Scenarios.} We construct three \textbf{generalization} test suites from RoboCasa assets: (1) \textit{Unseen Appearance (18 Tasks)}, novel visual textures on familiar container and object pairs; (2) \textit{Unseen Combinations (23 Tasks)}, seen objects in novel container pairings (14 pick-and-place, 9 articulated); (3) \textit{Unseen Object Types (32 Tasks)}, novel object categories.

\subsubsection{DROID Dataset}

The DROID dataset~\cite{khazatsky2024droid} contains 95,599 diverse trajectories collected from 564 scenes, including approximately 76k successful and 19k failed trajectories. The diversity of actions and scenes provides a comprehensive testbed for evaluating action-conditioned world models.

\subsubsection{Real-World Experiments}
\label{sec:exp_setup_real}

\begin{figure}[!t]
    \centering
    \includegraphics[width=0.95\textwidth]{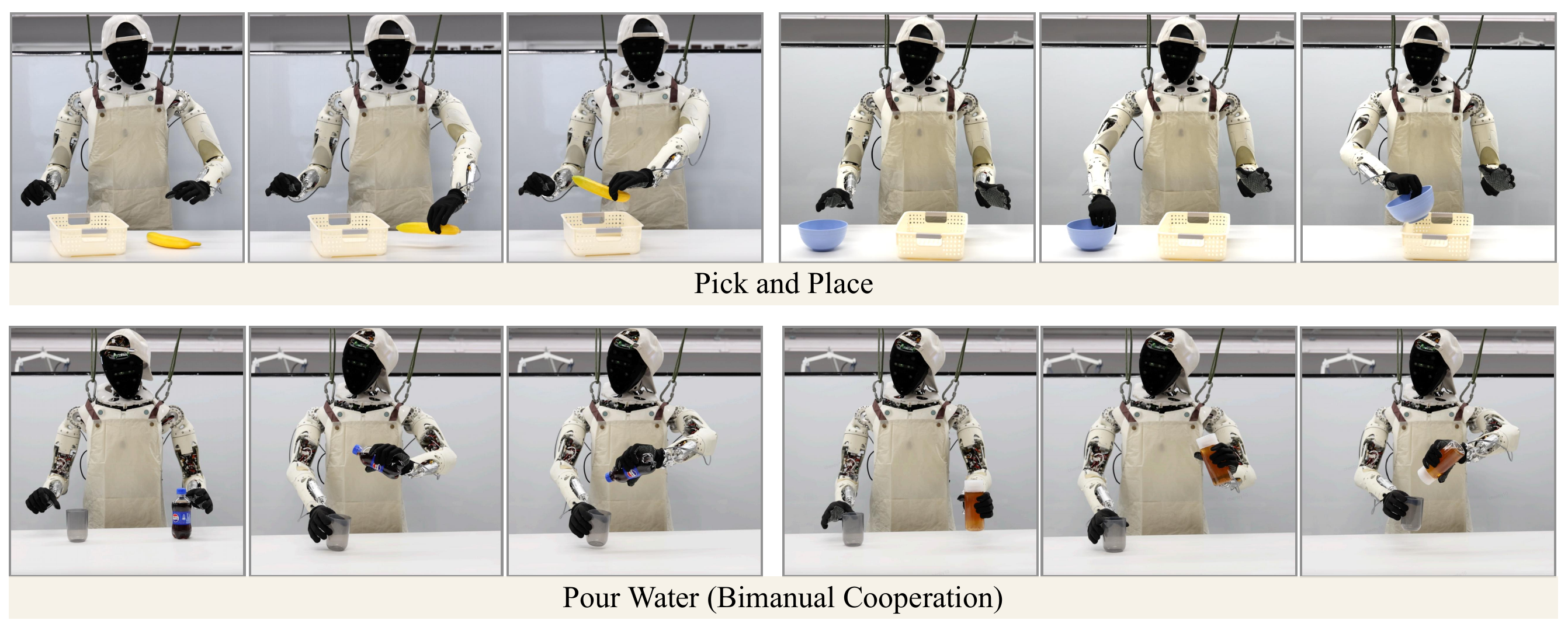}
    \caption{
    \textbf{Real-world in-domain tasks.} We design two tasks analogous to EgoDex subsets: \textbf{Pick \& Place} (pick an object and place it into a box) and \textbf{Pouring} (grasp a bottle and a cup, then pour).
    }
    \label{fig:real_robot_indomain}
\end{figure}

We validate on the IRON-R01-1.11 humanoid with a 50-dimensional action space covering arms, hands, waist, head, and wrist poses. As shown in Fig.~\ref{fig:real_robot_indomain}, we design two real-world tasks corresponding to EgoDex subsets: \textbf{Pick \& Place} (analogous to \texttt{basic\_pick\_place}, 27,419 trajectories) requires picking an object and placing it into a box, and \textbf{Pouring} (analogous to \texttt{pour}, 3,205 trajectories) requires bimanual grasping and pouring. We collect 120 robot trajectories per task. All models are pre-trained on a mixture of 32k proprietary robot trajectories and 30k EgoDex trajectories, then fine-tuned on task-specific data.



\begin{figure}[!t]
    \centering
    \includegraphics[width=0.95\textwidth]{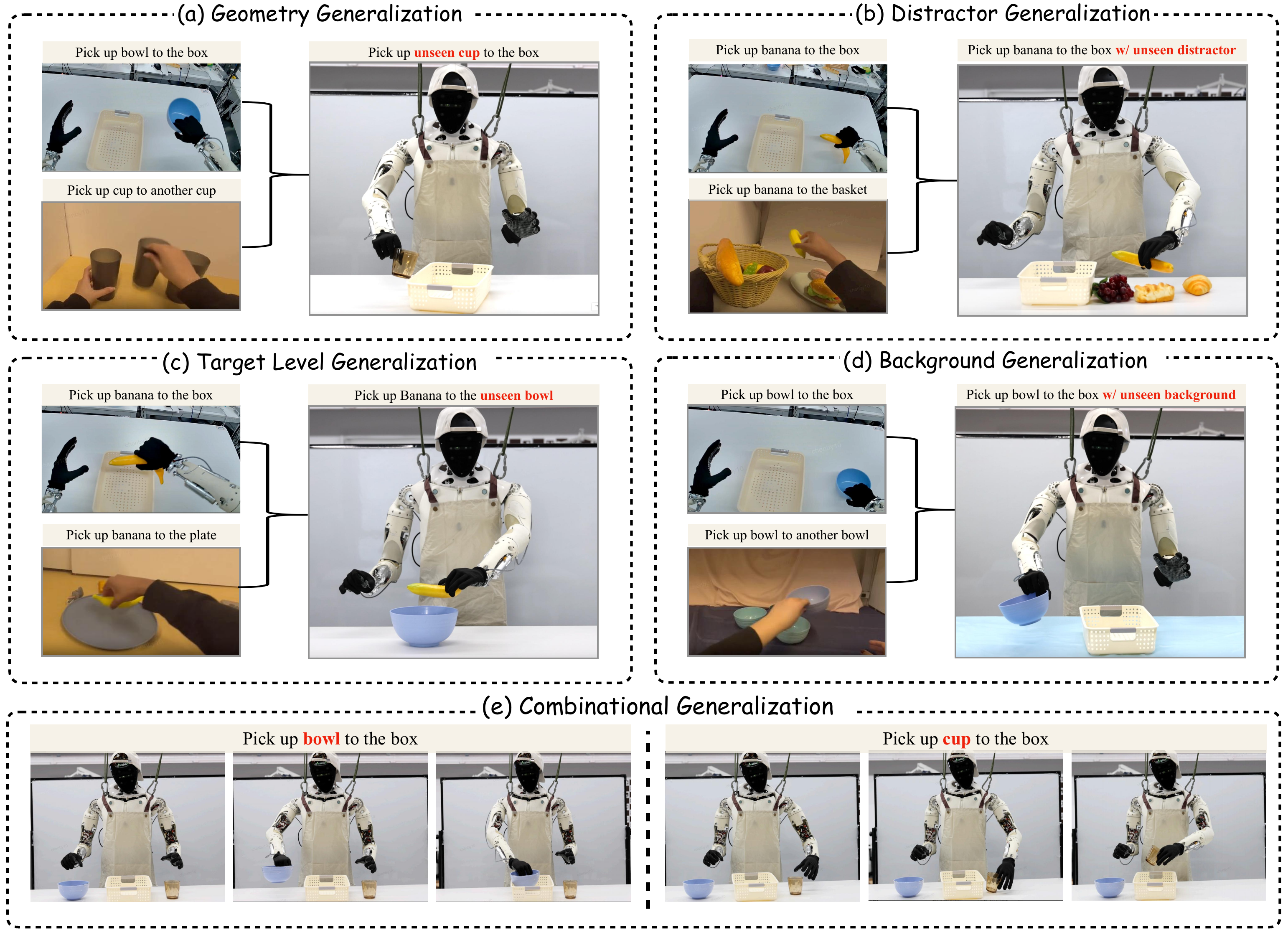}
    \caption{
    \textbf{Real-world OOD generalization scenarios.} \textbf{(a)--(d)}: human demonstrations introduce complementary variations absent from robot data; we co-train and evaluate on human-introduced conditions. \textbf{(a)} Geometry (new shapes, same affordance). \textbf{(b)} Target (new placement destinations). \textbf{(c)} Distractor (unseen objects introduced as distractors). \textbf{(d)} Visual (unseen surfaces). \textbf{(e)} Combinational (instruction-based disambiguation among multiple objects seen during training).
    }
    \label{fig:real_robot_ood}
\end{figure}

\paragraph{Generalization Scenarios.}
We construct five OOD axes (Fig.~\ref{fig:real_robot_ood}). For the first four, robot teleoperation covers part of the relevant variation, while EgoDex human demonstrations provide complementary conditions not sufficiently covered in robot data. We therefore co-train on both sources and evaluate on the human-introduced conditions.
\textit{(a) Geometry:} human data introduce the same affordance with new 3D shapes, such as cups of varying diameters in place of a bowl.
\textit{(b) Distractor:} human demonstrations include additional objects around the target manipulation object, creating distractor-rich scenes.
\textit{(c) Target:} human data introduce alternative placement destinations while the manipulation object remains the same.
\textit{(d) Background:} human data vary the table texture or background surface appearance.
Beyond transfer, \textit{(e) Combinational} evaluates instruction following among multiple target manipulation objects that are all seen during training, requiring the robot to disambiguate the correct object from the instruction.

\subsection{Evaluated Variants}
\label{sec:exp_variants}

\paragraph{Policy Variants and Baselines.}
In addition to \textbf{VLA-UniT} (our full model), we consider a \textbf{GR00T baseline} denoted as GR00T-Qwen2.5, which uses the same architecture as VLA-UniT, but without UniT token prediction. We further evaluate three tokenizer ablation variants corresponding to the paradigms in Fig.~\ref{fig:related_work}. All variants share the same VLA architecture and differ only in the token prediction target:
\textbf{VLA-Action} predicts action-only latent tokens without visual anchoring;
\textbf{VLA-Vision} predicts vision-only latent tokens without motor information;
and \textbf{VLA-UniT w/o Cross-Recon} predicts tokens from a tokenizer that encodes both modalities but without the cross-reconstruction objective, treating vision and action as decoupled vocabularies.
We additionally compare \textbf{VLA-Villa}, which replaces UniT with a tokenizer re-implemented following the design of Villa-X~\cite{chen2025villa} on our codebase, adopting a unidirectional vision-to-action (V2A) reconstruction objective rather than bidirectional cross-reconstruction.
For policy learning, we compare VLA-UniT against representative methods spanning diffusion-based control, flow-matching policies, and action-token prediction paradigms:
\textbf{Diffusion Policy}~\cite{chi2023diffusion}, which models actions via U-Net denoising;
\textbf{UWM}~\cite{li2025unified}, a transformer unifying action and video diffusion;
\textbf{FLARE}~\cite{zheng2025flare}, a flow-matching framework with future latent alignment;
\textbf{GR00T-N1.6}~\cite{gear2025gr00tn16}, an upgraded GR00T variant with a larger DiT backbone;
\textbf{GR00T-Qwen3}~\cite{gr00tn1_2025}, combining a frozen VLM with flow-matching action generation;
\textbf{$\pi$-Qwen3}~\cite{black2024pi_0}, coupling per-layer VLM features with a flow-matching expert;
\textbf{FAST-Qwen3}~\cite{pertsch2025fast}, using frequency-based BPE tokenization for autoregressive prediction;
and \textbf{OFT-Qwen3}~\cite{kim2025fine}, an optimized fine-tuning recipe with parallel action-chunk decoding.

\paragraph{World Modeling Variants and Metrics.}
For world modeling, we compare three conditioning paradigms: \textbf{Raw Action}, which conditions the world model on embodiment-specific raw actions; \textbf{WM-Action}, which uses action-only latent tokens without visual anchoring; and \textbf{WM-UniT}, which uses UniT's visually-anchored tokens (Sec.~\ref{wm-unit}). We evaluate generation quality with PSNR, SSIM, LPIPS~\cite{zhang2018unreasonable}, FVD~\cite{unterthiner2019fvd}, and EPE. These metrics capture frame fidelity, perceptual similarity, video realism, and controllability, respectively. Here, EPE denotes End-Point Error computed from optical flow.

\section{Experiments}
\label{sec:experiments}

We validate UniT through the following questions:
\begin{itemize}[nosep,leftmargin=*]
    \item \textbf{Q1 (Unified Representation):} Does UniT establish a shared physical language that aligns heterogeneous embodiments and remains robust to noise? (Sec.~\ref{sec:exp_repr})
    \item \textbf{Q2 (Efficient Policy Learning):} Does UniT enable data-efficient and generalizable policy learning for humanoids? (Sec.~\ref{sec:exp_vla_perf},~\ref{sec:exp_vla_human},~\ref{sec:exp_vla_real})
    \item \textbf{Q3 (Effective World Modeling):} Does UniT conditioning improve action-conditioned generation and enable cross-embodiment dynamics transfer? (Sec.~\ref{sec:exp_wm_control},~\ref{sec:exp_wm_transfer})
    \item \textbf{Q4 (Design Soundness):} How does UniT's visual-anchored design compare to alternative tokenizer paradigms? (Sec.~\ref{sec:exp_ablation})
\end{itemize}

\subsection{Unified Representation: Alignment and Robustness}
\label{sec:exp_repr}

\begin{figure}[!t]
    \centering
    \includegraphics[width=1.0\textwidth]{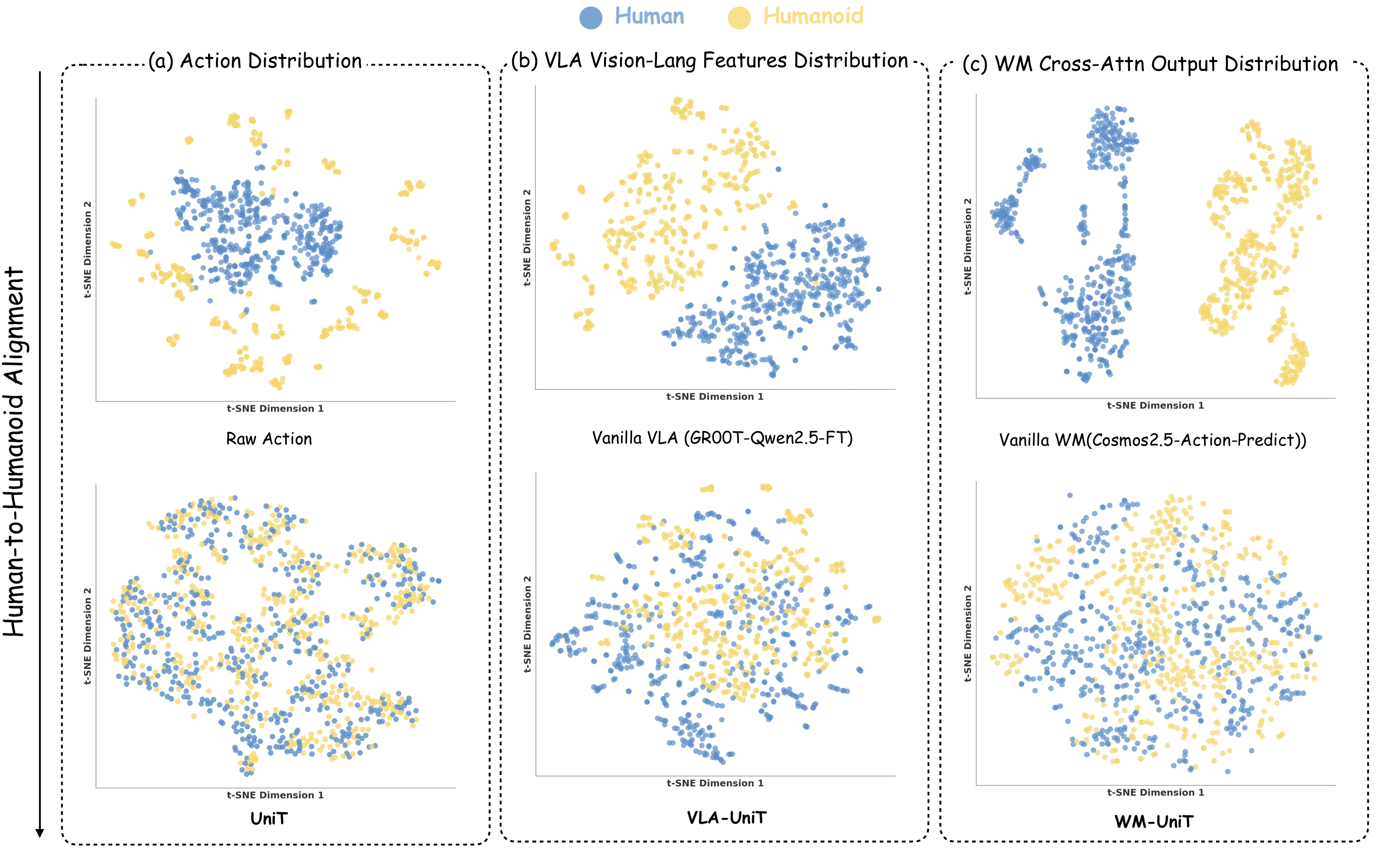}
    \caption{
    \textbf{Cross-embodiment representation alignment.} We plot t-SNE of human (blue) and humanoid (yellow) data across three levels: \textbf{(a)} raw actions vs.\ UniT token embeddings, \textbf{(b)} mean-pooled VLA vision-language hidden states, \textbf{(c)} mean-pooled WM cross-attention context embeddings. Vanilla baselines (top) show clearly separated distributions, while UniT-based models (bottom) produce highly overlapping representations, confirming that visual-anchored tokenization induces cross-embodiment alignment from the token level all the way to the internals of downstream models.
    }
    \label{fig:alignment}
\end{figure}


We first examine whether UniT establishes the \textit{unified physical language} claimed in Sec.~\ref{sec:unit}, and whether this alignment propagates into downstream models. We perform t-SNE~\cite{JMLR:v9:vandermaaten08a} analysis on human and humanoid samples from the RoboCasa GR1 and EgoDex co-training mixture (Fig.~\ref{fig:alignment}). For downstream VLA, we compare against GR00T-Qwen2.5-FT, which uses Qwen2.5-VL as backbone and fine-tunes the core language modeling blocks to predict raw action. For downstream world modeling, we compare against Cosmos Predict 2.5 with raw action conditioning. Both baselines are trained on the same human-humanoid data mixture.

\paragraph{Cross-Embodiment Token Alignment.}
We compare the distributions of raw action trajectories and UniT token embeddings (Fig.~\ref{fig:alignment}a). In the raw action space, human and humanoid data form clearly separated clusters, reflecting the inherent distribution gap between heterogeneous kinematics. After encoding through UniT, the token embeddings become highly overlapping, confirming that the visual-anchored cross-reconstruction (Sec.~\ref{sec:unit}) successfully projects disparate action spaces into a shared manifold.


\begin{figure}[!t]
    \centering
    \includegraphics[width=1.0\textwidth]{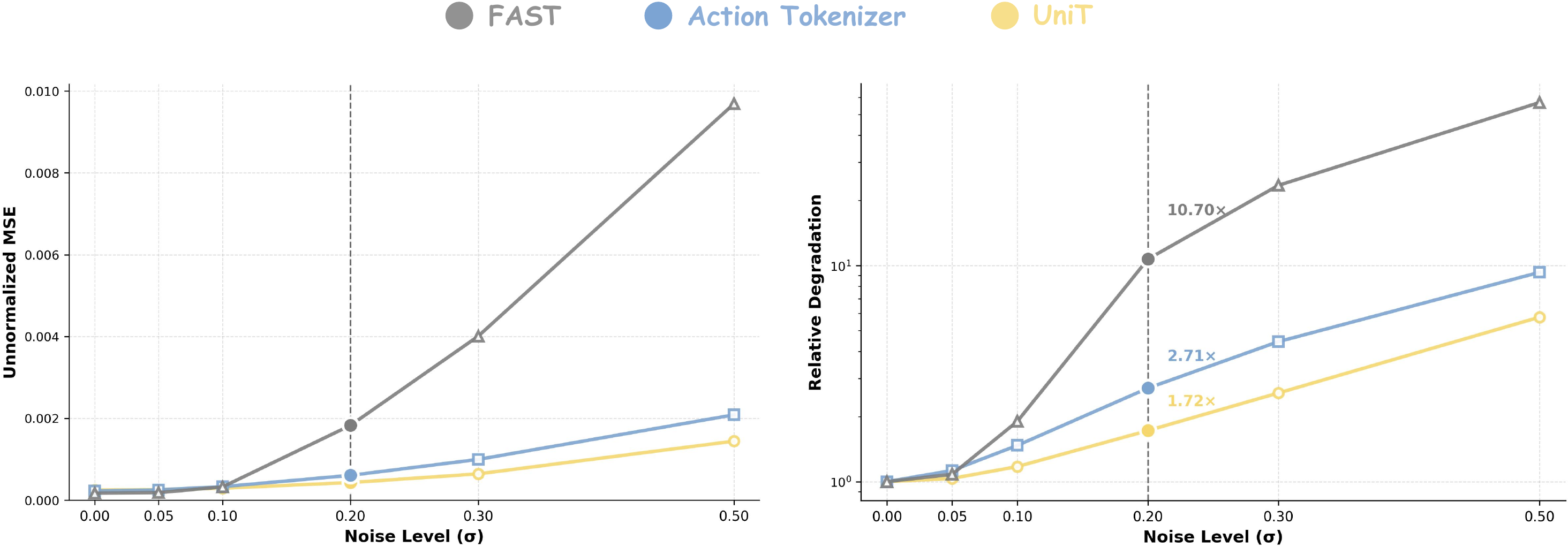}
    \caption{
    \textbf{Robustness to action noise.} Gaussian noise of varying intensity $\sigma$ is injected into EgoDex action trajectories. Here, $\sigma$ denotes the relative noise level normalized by the global action standard deviation of the dataset, so the injected noise magnitude is $\sigma \times \text{global std}$. Reconstruction quality is evaluated using mean squared error (MSE) between reconstructed actions and the original clean actions. \textbf{(Left)} Absolute (unnormalized) MSE, zoomed in to $\sigma \leq 0.5$. \textbf{(Right)} Relative degradation, defined as $\text{MSE}_{\text{noisy}} / \text{MSE}_{\text{clean}}$, shown on a log scale. At $\sigma = 0.2$ (vertical reference line), FAST degrades by $10.7\times$, the action-only tokenizer by $2.7\times$, and UniT by only $1.7\times$, showing that visual grounding in UniT provides effective denoising.
    }
    \label{fig:noise_robustness}
\end{figure}

\paragraph{Robustness to Action Noise.}
In-the-wild human motion capture data inevitably contains noise from sensor jitter and annotation artifacts. We evaluate whether the cross-reconstruction mechanism in UniT (Sec.~\ref{sec:unit}), which grounds actions in visual transitions, provides implicit denoising. We inject Gaussian noise of varying intensity $\sigma$ into EgoDex action trajectories, where $\sigma$ denotes the relative noise level normalized by the global action standard deviation of the dataset, and compare the reconstruction quality of three tokenizers: FAST~\cite{pertsch2025fast}, a frequency-based BPE action tokenizer; Action Tokenizer~\cite{lee2024behavior}, which uses the same RQ-VAE architecture as UniT but is trained on action data alone; and UniT, which jointly leverages visual and action information. Reconstruction quality is measured by mean squared error (MSE) between reconstructed actions and the original clean actions. As shown in Fig.~\ref{fig:noise_robustness}, UniT exhibits the lowest absolute MSE (left) and the lowest relative degradation $\text{MSE}_{\text{noisy}} / \text{MSE}_{\text{clean}}$ (right) across noise levels. At $\sigma = 0.2$, FAST degrades by $10.7\times$ and the action-only tokenizer by $2.7\times$, while UniT degrades by only $1.7\times$, maintaining near-clean reconstruction quality. The gap between UniT and the action-only tokenizer confirms that visual grounding provides complementary information that regularizes the latent space, filtering out kinematic noise that lacks visual correspondence and yielding more robust representations for downstream deployment.

\paragraph{Downstream Representation Alignment.}

We further examine whether token-level alignment propagates into downstream model internals.For VLA, we extract mean-pooled last-layer vision-language features after the action expert self-attention; for WM, we extract mean-pooled last-layer cross-attention outputs after UniT feature injection (Sec.~\ref{wm-unit}). The vanilla VLA (GR00T-Qwen2.5VL-FT, an LLM-tuned GR00T variant based on Qwen2.5-VL without UniT token prediction supervision) and vanilla WM (Cosmos Predict 2.5) using raw actions as conditioning follow the same architectures introduced in Sec.~\ref{sec:vla-unit} and Sec.~\ref{wm-unit}. As shown in Fig.~\ref{fig:alignment}(b), the vanilla VLA maintains clearly separated human and humanoid feature distributions, while VLA-UniT produces substantially more interleaved representations. The effect is even more pronounced in world modeling (Fig.~\ref{fig:alignment}c): the vanilla WM exhibits fully disjoint clusters, whereas WM-UniT brings them into a single unified distribution. These results confirm that UniT tokens not only form a shared cross-embodiment latent space, but also induce downstream models to develop embodiment-agnostic internal representations, providing a structural basis for the performance gains observed in the following sections.

\subsection{Policy Learning}

\subsubsection{Benchmark Performance and Data Efficiency}
\label{sec:exp_vla_perf}

As established in Sec.~\ref{sec:vla-unit}, UniT token prediction provides the VLM with a compact, visually-anchored learning objective that encodes physical intent. We evaluate whether this translates to improved policy performance and data efficiency on the RoboCasa GR1 benchmark, compared against the policy baselines described in Sec.~\ref{sec:exp_variants}.


\begin{figure}[!t]
    \centering
    \includegraphics[width=1.0\textwidth]{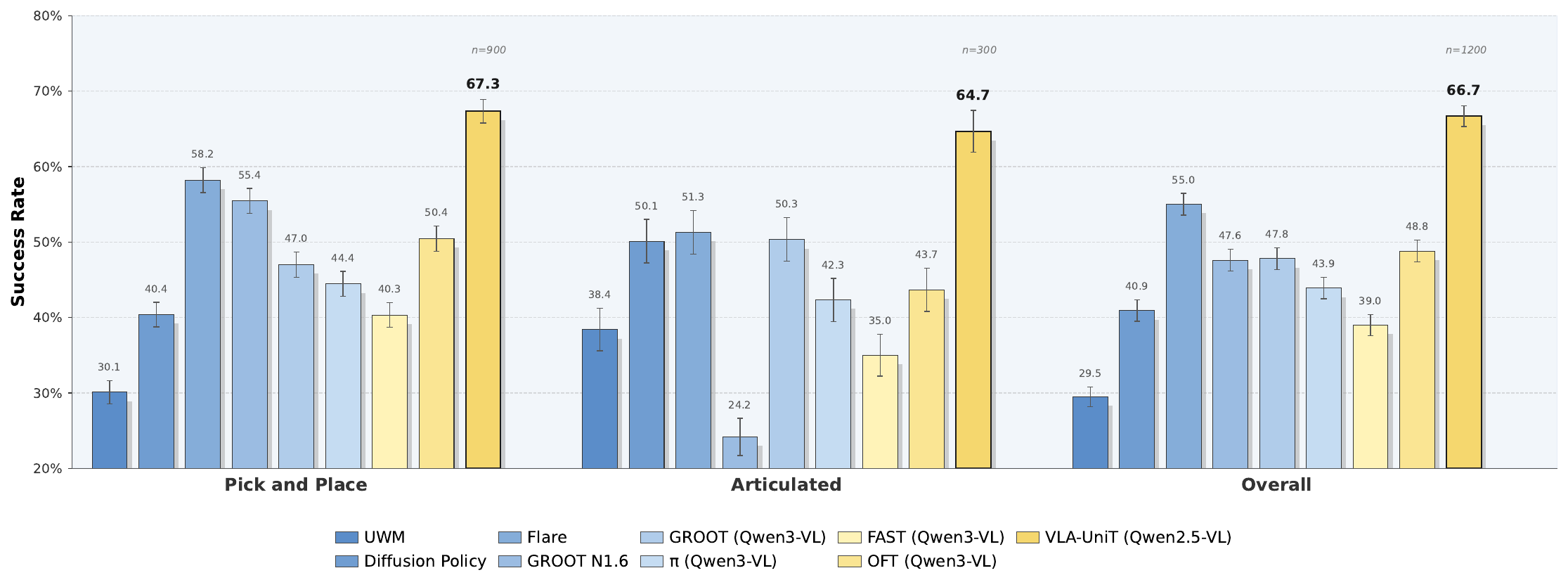}
    \caption{
    \textbf{Overall performance} on RoboCasa GR1 Tabletop Simulation with full training data. VLA-UniT achieves the strongest overall policy performance among all compared methods.
    }
    \label{fig:method_comparison}
\end{figure}

\paragraph{Overall Performance.} As shown in Fig.~\ref{fig:method_comparison}, VLA-UniT achieves a 66.7\% overall success rate on the full-data RoboCasa benchmark, outperforming all baselines by a substantial margin. In Pick \& Place tasks, which require grounding semantic instructions into precise object rearrangement behaviors, VLA-UniT attains 67.3\%. In Articulated tasks, where the robot must manipulate articulated fixtures such as cabinets and microwaves through multi-step interactions, VLA-UniT reaches 64.7\%, maintaining consistently strong performance across both categories. VLA-UniT surpasses all baselines, outperforming the previous best FLARE (55.0\%) by 11.7\%. Notably, compared to the GR00T baseline (47.8\%), which shares the same architecture without UniT token prediction, the improvement of 18.9\% highlights the value of introducing UniT as a learning objective (Sec.~\ref{sec:vla-unit}). By jointly grounding visual transitions and motor commands through cross-reconstruction (Sec.~\ref{sec:unit}), UniT tokens provide the VLM with a compact prediction target that strengthens visuo-motor synergy, enabling more effective policy learning across diverse task types.

\begin{figure}[!t]
    \centering
    \includegraphics[width=0.34\textwidth]{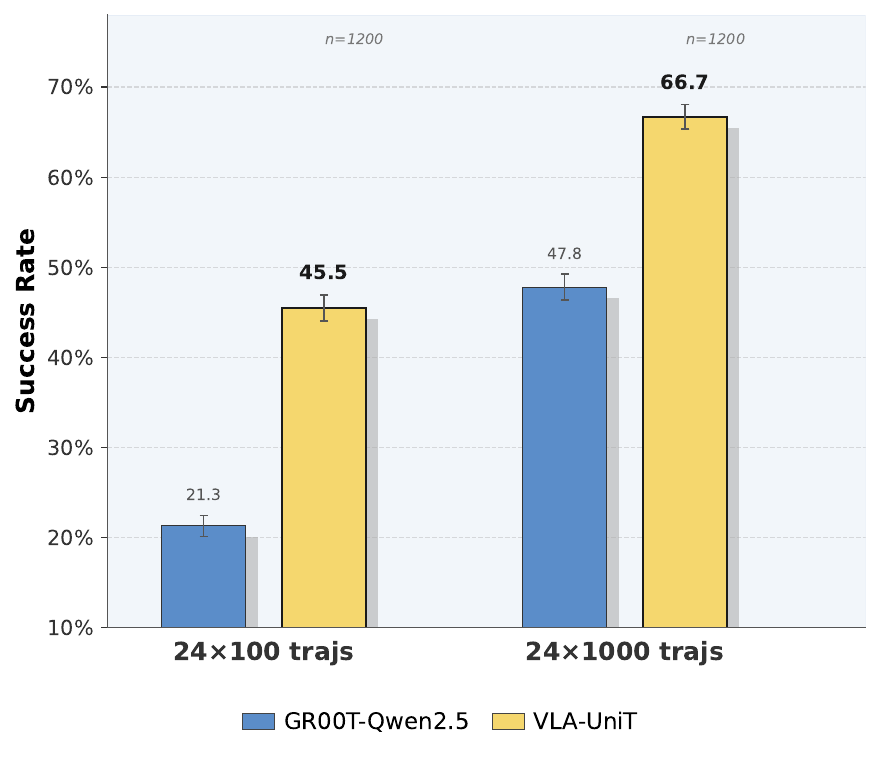}
    \hfill
    \includegraphics[width=0.62\textwidth]{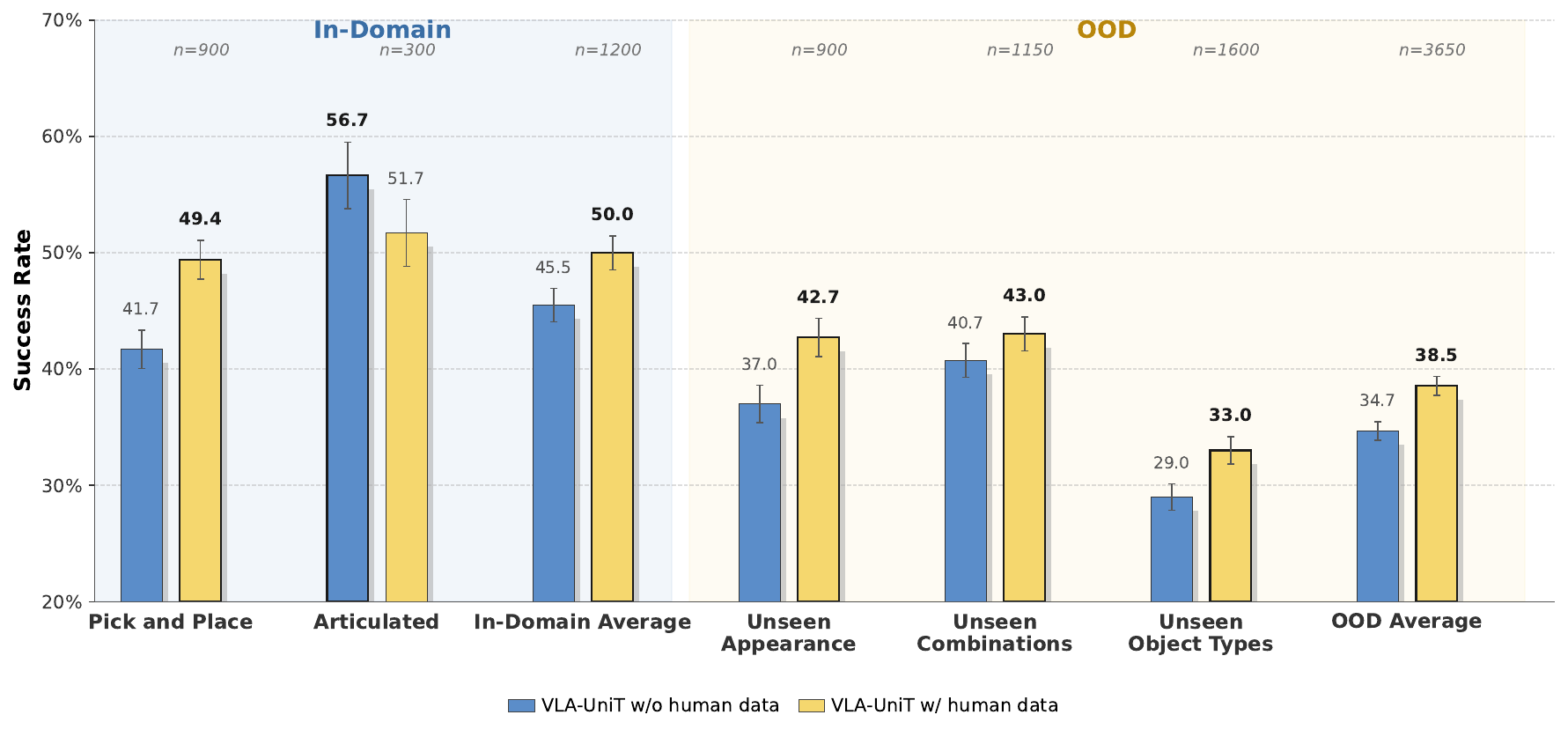}
    \caption{
    \textbf{(Left)} Data efficiency results on RoboCasa GR1 under full-data and few-shot settings. \textbf{(Right)} Impact of incorporating EgoDex  \texttt{basic\_pick\_place} human demonstrations on few-shot performance in RoboCasa GR1. UniT improves both sample efficiency and the utility of human co-training.
    }
    \label{fig:data_efficiency}
\end{figure}

\paragraph{Data Efficiency.} To assess sample efficiency, we compare VLA-UniT and the GR00T baseline under both full-data and few-shot regimes (Fig.~\ref{fig:data_efficiency}, left). With only 10\% of the training data (100 trajectories per task), VLA-UniT achieves 45.5\% success rate, already approaching the GR00T baseline trained on full data (47.8\%). This $\sim$10$\times$ reduction in data requirements highlights the advantage of UniT's compact, cross-modality aligned representation: by operating in a structured discrete latent space rather than regressing raw actions, the VLM extracts task-relevant intent more efficiently from limited demonstrations.

\subsubsection{Human-to-Humanoid Transfer}
\label{sec:exp_vla_human}

We investigate whether UniT's shared latent space enables leveraging large-scale human demonstrations to improve humanoid policy learning. Under the few-shot regime in simulation, we co-train VLA-UniT on robot data and EgoDex human demonstrations, then fine-tune on robot data alone (Fig.~\ref{fig:data_efficiency}, right).

Incorporating human data improves both in-domain and OOD performance. The in-domain average performance increases from 45.5\% to 50.0\%, with the largest gain in Pick \& Place (41.7\% $\to$ 49.4\%), which directly corresponds to the EgoDex \texttt{basic\_pick\_place} domain. Across all three OOD scenarios, human co-training brings consistent improvements: Unseen Appearance (37.0\% $\to$ 42.7\%), Unseen Combinations (40.7\% $\to$ 43.0\%), and Unseen Object Types (29.0\% $\to$ 33.0\%), yielding an OOD average gain from 34.7\% to 38.5\%. These results confirm that UniT's shared latent space enables effective human-to-humanoid transfer. As shown in Sec.~\ref{sec:exp_repr}, VLA-UniT produces highly overlapping vision-language representations for human and humanoid data (Fig.~\ref{fig:alignment}b), indicating that the VLM builds a unified internal feature space across embodiments. This representational alignment provides the structural basis for transfer: human demonstrations, mapped into the same token vocabulary, broaden the VLM's task coverage and directly benefit humanoid policy generalization even in out-of-distribution settings.

\subsubsection{Real-World Generalization}
\label{sec:exp_vla_real}

\begin{figure}[!t]
    \centering
    \includegraphics[width=0.34\textwidth]{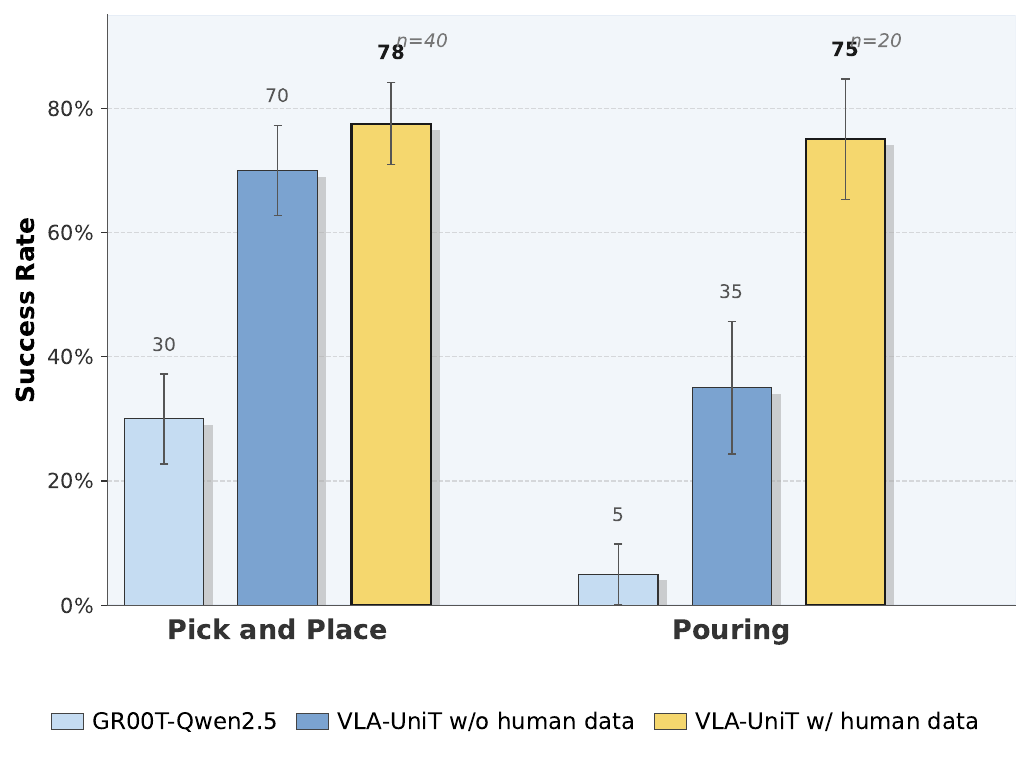}
    \hfill
    \includegraphics[width=0.62\textwidth]{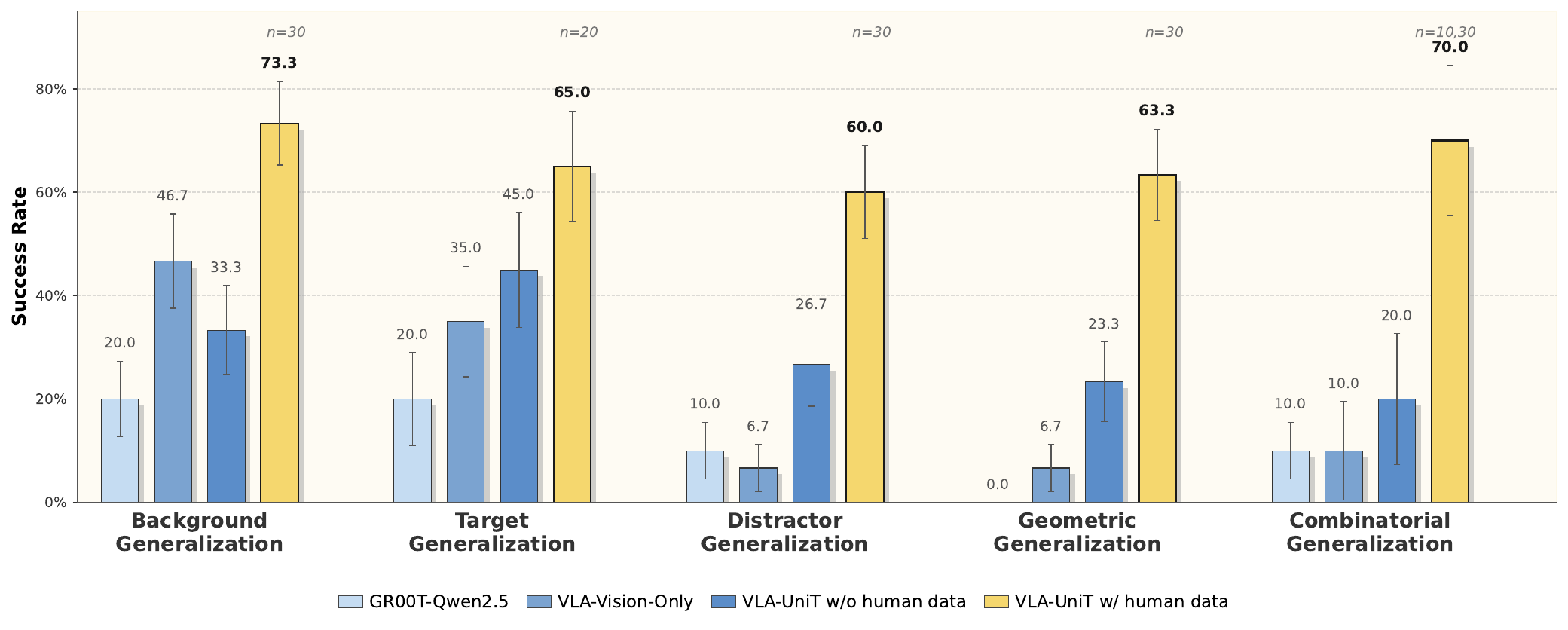}
    \caption{
    \textbf{(Left)} In-domain results on the IRON-R01-1.11 robot. \textbf{(Right)} OOD generalization results on the IRON-R01-1.11 robot. Human Demonstration improves VLA-UniT's real-world execution and OOD robustness.
    }
    \label{fig:real_world}
\end{figure}


We deploy VLA-UniT on the real-world IRON-R01-1.11 humanoid to validate whether the performance, human-to-humanoid transfer, and generalization gains observed in simulation also carry over to physical deployment.

\paragraph{In-Domain Performance.}
We evaluate VLA-UniT on two real-world tasks (Fig.~\ref{fig:real_world}, left). With robot data alone, VLA-UniT already substantially outperforms the GR00T baseline on both Pick \& Place (70\% vs.\ 30\%) and Pouring (35\% vs.\ 5\%). Incorporating EgoDex human data further improves performance to 78\% and 75\% respectively, with the gain particularly pronounced in Pouring, a task requiring coordinated dual-arm control that benefits from the rich bimanual interaction patterns in human demonstrations. UniT's shared latent space enables the VLM to directly leverage human bimanual coordination experience for humanoid execution.

\paragraph{OOD Generalization.}
As described in Sec.~\ref{sec:exp_setup_real}, the first four OOD axes are designed so that robot data provide partial coverage while human demonstrations introduce the complementary variation; we evaluate on the human-introduced conditions. Across all five categories (Fig.~\ref{fig:real_world}, right), VLA-UniT with human co-training consistently achieves the strongest performance. In Geometry and Distractor Generalization scenarios — where human videos introduce novel object shapes and visual clutter respectively — the improvement is most pronounced (23.3\%$\to$63.3\% and 26.7\%$\to$60.0\%), confirming that human data effectively fills the variation gap left by limited robot demonstrations. Background and Target-Level Generalizations show similar trends, indicating that the relevant visual and goal-conditioned knowledge also transfers through UniT's shared space. Notably, VLA-Vision shows limited OOD robustness in these scenarios, confirming the limitations of vision-only representations discussed in Sec.~\ref{sec:unit}. The Combinational setting, which tests language-guided disambiguation, further improves from 10\% to 70\%, suggesting that the broader interaction diversity from human co-training also strengthens compositional generalization.

\begin{figure}[!t]
    \centering
    \includegraphics[width=1.0\textwidth]{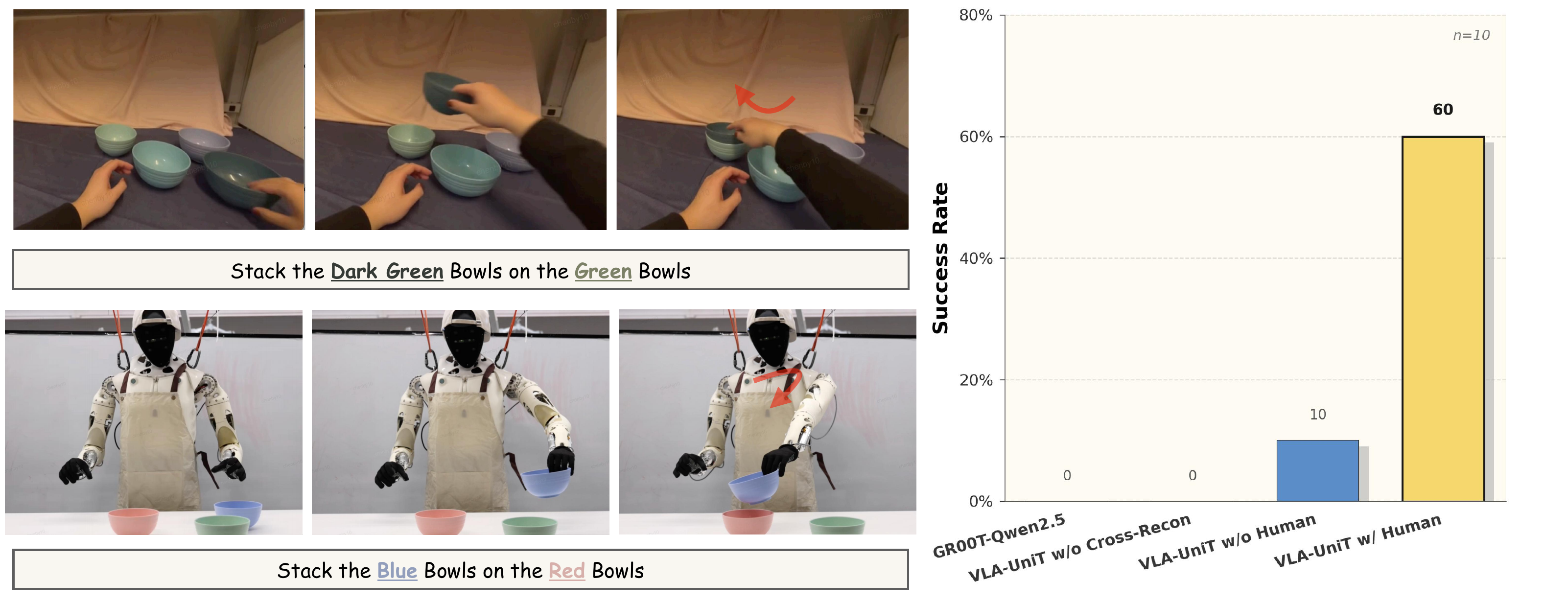}
    \caption{
    \textbf{Zero-shot task transfer} on the IRON-R01-1.11 robot. (Left) Task illustration. (Right) Success rates on the unseen stacking task. UniT with human co-training shows the clearest transfer to the unseen task.
    }
    \label{fig:zero_shot}
\end{figure}

\paragraph{Zero-Shot Task Transfer.}
We evaluate on a stacking task that is not covered by robot training demonstrations (Fig.~\ref{fig:zero_shot}). Robot data only include pick-and-place of individual bowls, while EgoDex human videos contain stacking sequences performed with view switching and upper-body coordination. The GR00T baseline and VLA-UniT without cross-reconstruction both score 0\%, indicating that neither raw action fitting nor tokenization without cross-modal alignment can bridge the task gap. VLA-UniT without human data achieves 10\%. With human co-training, VLA-UniT reaches 60\%, transferring the stacking logic from human demonstrations and exhibiting emergent upper-body coordination — waist rotation and head turning to adjust viewpoint — that mirrors the coordination patterns observed in human videos. This demonstrates that UniT's visual-anchored cross-reconstruction creates a representational bridge strong enough to transfer not just task semantics but also fine-grained coordination patterns across embodiments.

\subsection{World Modeling}

\subsubsection{Controllable Generation}
\label{sec:exp_wm_control}

\begin{table}[t]
  \caption{\textbf{Controllable generation} results on DROID and EgoDex + RoboCasa-GR1 co-training.}
  \label{tab:wm-comparison}
  \centering
  \small
  \begin{tabular}{llccccc}
    \toprule
    Dataset & Method & PSNR $\uparrow$ & SSIM $\uparrow$ & LPIPS $\downarrow$ & FVD $\downarrow$ & EPE $\downarrow$ \\
    \midrule
    \multicolumn{7}{l}{\textit{Single-embodiment (DROID)}} \\
    \multirow{3}{*}{DROID}
    & Raw Action    & \underline{21.02} & \underline{0.820} & \underline{0.097} & \textbf{76.38} & 0.2662 \\
    & WM-Action & 20.86 & 0.819 & 0.102 & 80.30 & \underline{0.2593} \\
    & WM-UniT & \textbf{21.32} & \textbf{0.823} & \textbf{0.095} & \underline{76.44} & \textbf{0.2588} \\
    \midrule
    \multicolumn{7}{l}{\textit{Human-humanoid co-training (EgoDex + RoboCasa-GR1)}} \\
    \multirow{2}{*}{EgoDex}
    & Raw Action    & 24.84 & 0.800 & 0.164 & 171.37 & 0.706 \\
    & WM-UniT & \textbf{28.06} & \textbf{0.858} & \textbf{0.086} & \textbf{130.87} & \textbf{0.519} \\
    \multirow{2}{*}{RoboCasa-GR1}
    & Raw Action    & 13.45 & 0.590 & 0.259 & 237.13 & 0.558 \\
    & WM-UniT & \textbf{17.66} & \textbf{0.718} & \textbf{0.142} & \textbf{166.50} & \textbf{0.453} \\
    \bottomrule
  \end{tabular}
\end{table}

As described in Sec.~\ref{wm-unit}, WM-UniT conditions video generation on UniT's continuous pre-quantization features. We then examine whether this conditioning interface improves controllability and cross-embodiment transfer under a fixed video generation backbone. In all world-model experiments, we evaluate 10-step autoregressive rollouts over 10-second video generation; DROID is evaluated at resolution $192\times320$, while RoboCasa-GR1 and EgoDex are evaluated at $192\times336$. We compare UniT Tokens against raw actions and action-only latent tokens. 



As shown in Table~\ref{tab:wm-comparison}, WM-UniT achieves the strongest controllability on DROID, as reflected by the best EPE. It also improves PSNR, SSIM, and LPIPS over both baselines, while remaining competitive on FVD. In contrast, WM-Action does not yield a similarly reliable gain, indicating that latent tokenization alone is insufficient for controllable video generation. The results demonstrate that UniT provides a more compact and more faithful conditioning signal for world modeling.

\subsubsection{Human-Humanoid Transfer}
\label{sec:exp_wm_transfer}

\begin{table}[t]
  \caption{\textbf{Human data pre-training} for world modeling. Pre-trained on EgoDex \texttt{basic\_pick\_place}, fine-tuned on RoboCasa-GR1 pick-and-place. Human pre-training through UniT improves downstream humanoid controllability.}
  \label{tab:wm-human-pretrain}
  \centering
  \small
  \begin{tabular}{lccccc}
    \toprule
    Configuration & PSNR $\uparrow$ & SSIM $\uparrow$ & LPIPS $\downarrow$ & FVD $\downarrow$ & EPE $\downarrow$ \\
    \midrule
    WM-UniT w/o Human Pre-training & 16.34 & 0.678 & 0.168 & 180.51 & 0.478 \\
    WM-UniT (Full)             & \textbf{18.06} & \textbf{0.713} & \textbf{0.135} & \textbf{153.31} & \textbf{0.446} \\
    \bottomrule
  \end{tabular}
\end{table}

We further investigate whether UniT's shared latent space enables human-to-humanoid dynamics transfer for world modeling. As established in Sec.~\ref{sec:exp_repr}, WM-UniT produces unified context embeddings for human and humanoid data (Fig.~\ref{fig:alignment}c), suggesting that the world model builds a shared internal dynamics representation across embodiments.

\paragraph{Co-Training.}
Under joint training on EgoDex and RoboCasa-GR1 (Table~\ref{tab:wm-comparison}), WM-UniT consistently outperforms Raw Action on both the human and humanoid subsets. The gain is reflected not only in reconstruction quality, but more importantly in stronger controllability, indicating that UniT provides a shared conditioning space that supports cross-embodiment dynamics modeling. Combined with the aligned context embeddings in Fig.~\ref{fig:alignment}(c), this result shows that UniT enables the world model to co-train on human and humanoid data without collapsing into embodiment-specific dynamics.

\paragraph{Pre-Training.}
We pre-train WM-UniT on EgoDex \texttt{basic\_pick\_place} human demonstrations, then fine-tune on RoboCasa-GR1 pick-and-place data (Table~\ref{tab:wm-human-pretrain}). Human pre-training brings consistent gains across all metrics, with the most meaningful improvement reflected in controllability. This indicates that the dynamics learned from human data remain usable after transfer to humanoid prediction, rather than being tied to human-specific kinematics. Together with the shared representation analysis in Sec.~\ref{sec:exp_repr}, these results confirm that UniT provides a transferable dynamics interface for world modeling.

\paragraph{Cross-Embodiment Conditioning.}

\begin{figure}[!t]
    \centering
    \includegraphics[width=0.95\textwidth]{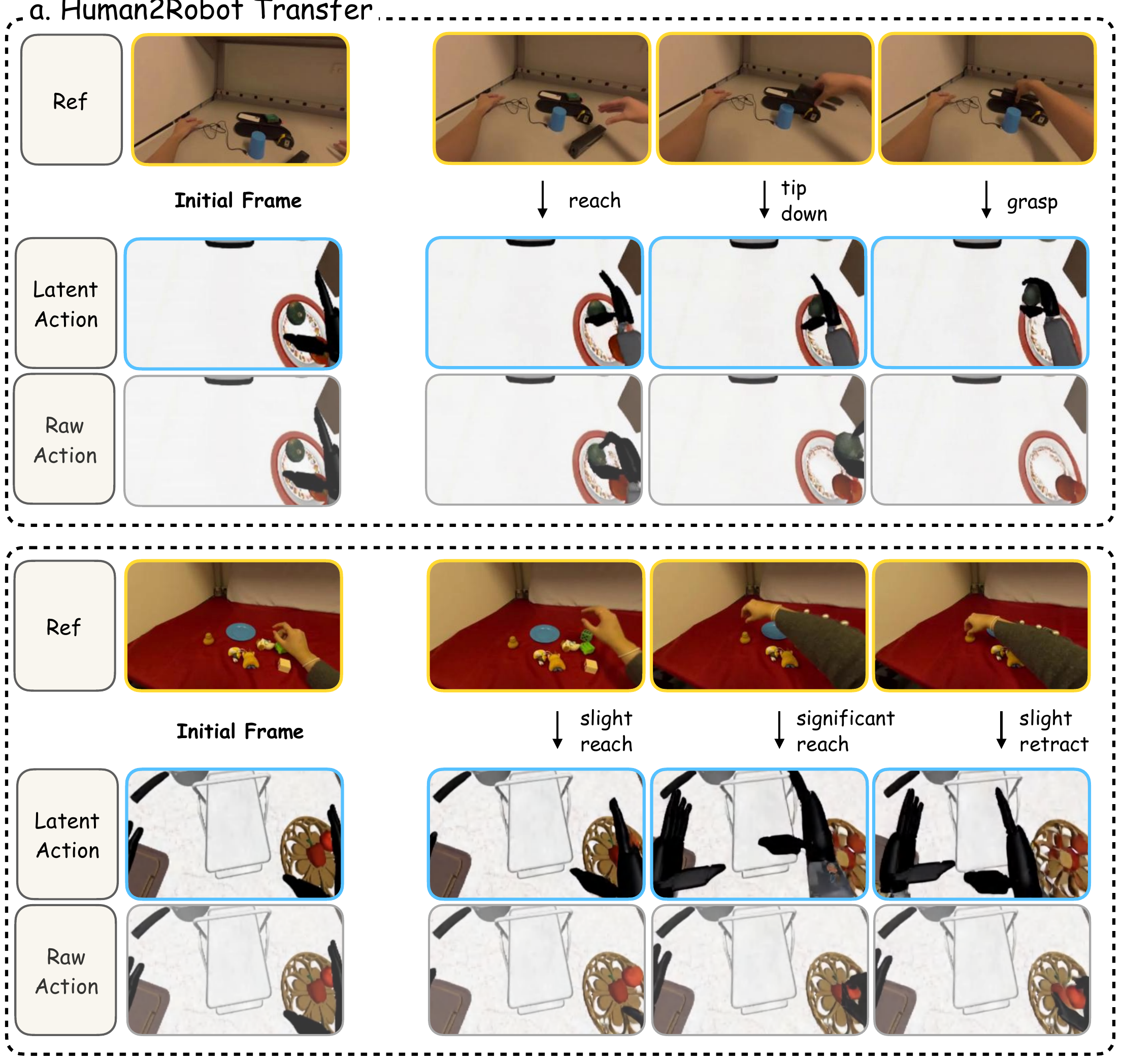}
    \caption{
    \textbf{Human-to-robot conditioning.} Top: human reference video. Bottom: human reference actions condition robot video generation via UniT vs.\ Raw Action. UniT yields more faithful cross-embodiment conditioning than raw actions.
    }
    \label{fig:wm_h2r}
\end{figure}

\begin{figure}[!t]
    \centering
    \includegraphics[width=0.95\textwidth]{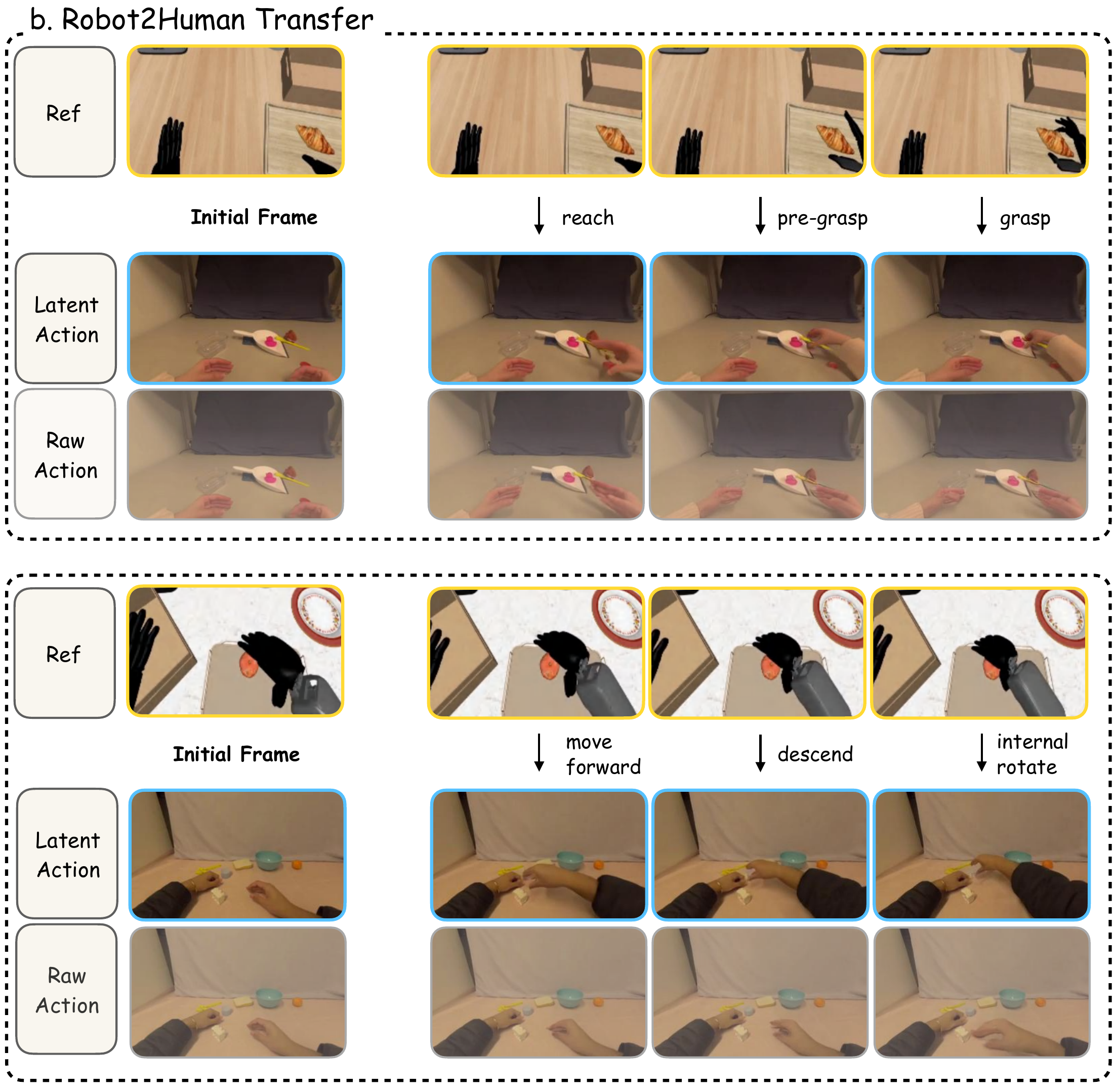}
    \caption{
    \textbf{Robot-to-human conditioning.} Top: robot reference video. Bottom: robot reference actions condition human video generation via UniT vs.\ Raw Action. UniT yields more faithful cross-embodiment conditioning than raw actions.
    }
    \label{fig:wm_r2h}
\end{figure}

Beyond co-training and pre-training, we directly test whether UniT tokens from one embodiment can condition video generation for the other — without any domain-specific adaptation. We condition the world model with actions from a source embodiment and generate videos for the target embodiment, comparing UniT against Raw Action conditioning (Fig.~\ref{fig:wm_r2h},~\ref{fig:wm_h2r}).



In the human-to-robot setting (Fig.~\ref{fig:wm_h2r}), the human reference includes a reach, tip-down, and grasp sequence, as well as varying magnitudes — a slight exploratory reach, a significant forward extension, and a slight retraction. UniT-conditioned generation distinguishes between these scales and preserves the non-monotonic trajectory (reach then retract), while Raw Action produces uniform motion that neither reflects magnitude differences nor captures directional reversals.

In the robot-to-human setting (Fig.~\ref{fig:wm_r2h}), the robot reference shows a multi-phase sequence — forward approach, vertical descent, and internal wrist rotation before grasping. UniT-conditioned generation faithfully reproduces each phase in the human domain, preserving both the grasp semantics and fine-grained pose adjustments such as the terminal rotation and tip-down motion. Raw Action conditioning captures the coarse trajectory but collapses these atomic actions into a flat reach, losing the rotational and postural details.

Together, these examples highlight three capabilities of UniT's cross-embodiment encoding: (1) \textit{fine-grained action semantics} — atomic actions such as internal rotation, tip-down, and grasp transfer across embodiments; (2) \textit{magnitude sensitivity} — the distinction between slight and significant motions is preserved; and (3) \textit{temporal coherence} — non-monotonic trajectories including retraction are faithfully reproduced.

To quantify these observations, we evaluate cross-embodiment conditioning consistency using Gemini-3-Pro as an automated judge. For each generated video paired with its reference, the model scores three dimensions on a 1--5 scale: \textit{Semantic} consistency (whether the intended action is preserved), \textit{Temporal} consistency (whether the motion timing and sequencing match), and \textit{Geometric} consistency (whether spatial trajectories and pose details are faithful).

\begin{table}[t]
  \caption{\textbf{Cross-embodiment conditioning consistency.} MLLM-based evaluation (Gemini-3-Pro) on EgoDex and RoboCasa-GR1. UniT improves semantic, temporal, and geometric consistency in both transfer directions.}
  \label{tab:wm-transfer-score}
  \centering
  \small
  \begin{tabular}{lcccc}
    \toprule
    Method & Semantic $\uparrow$ & Temporal $\uparrow$ & Geometric $\uparrow$ & Overall $\uparrow$ \\
    \midrule
    \multicolumn{5}{l}{\textit{Robot-to-Human}} \\
    Raw Action & 2.96 & 3.12 & 2.74 & 2.92 \\
    WM-UniT & \textbf{3.91} & \textbf{3.98} & \textbf{3.66} & \textbf{3.84} \\
    \midrule
    \multicolumn{5}{l}{\textit{Human-to-Robot}} \\
    Raw Action & 2.98 & 3.16 & 2.72 & 2.95 \\
    WM-UniT & \textbf{3.28} & \textbf{3.43} & \textbf{3.09} & \textbf{3.27} \\
    \bottomrule
  \end{tabular}
\end{table}

As shown in Table~\ref{tab:wm-transfer-score}, WM-UniT consistently outperforms Raw Action across all three evaluation dimensions in both directions. The Semantic score reflects the preservation of action intent (grasp, rotate, retract); the Temporal score captures sequencing fidelity and the ability to reproduce non-monotonic trajectories; and the Geometric score measures spatial precision including magnitude sensitivity and pose detail. WM-UniT's consistent improvement across all three confirms that it encodes a precise and transferable action representation for cross-embodiment world modeling.

\subsection{Tokenizer Design Ablation}
\label{sec:exp_ablation}

\begin{figure}[!t]
    \centering
    \includegraphics[width=0.61\textwidth]{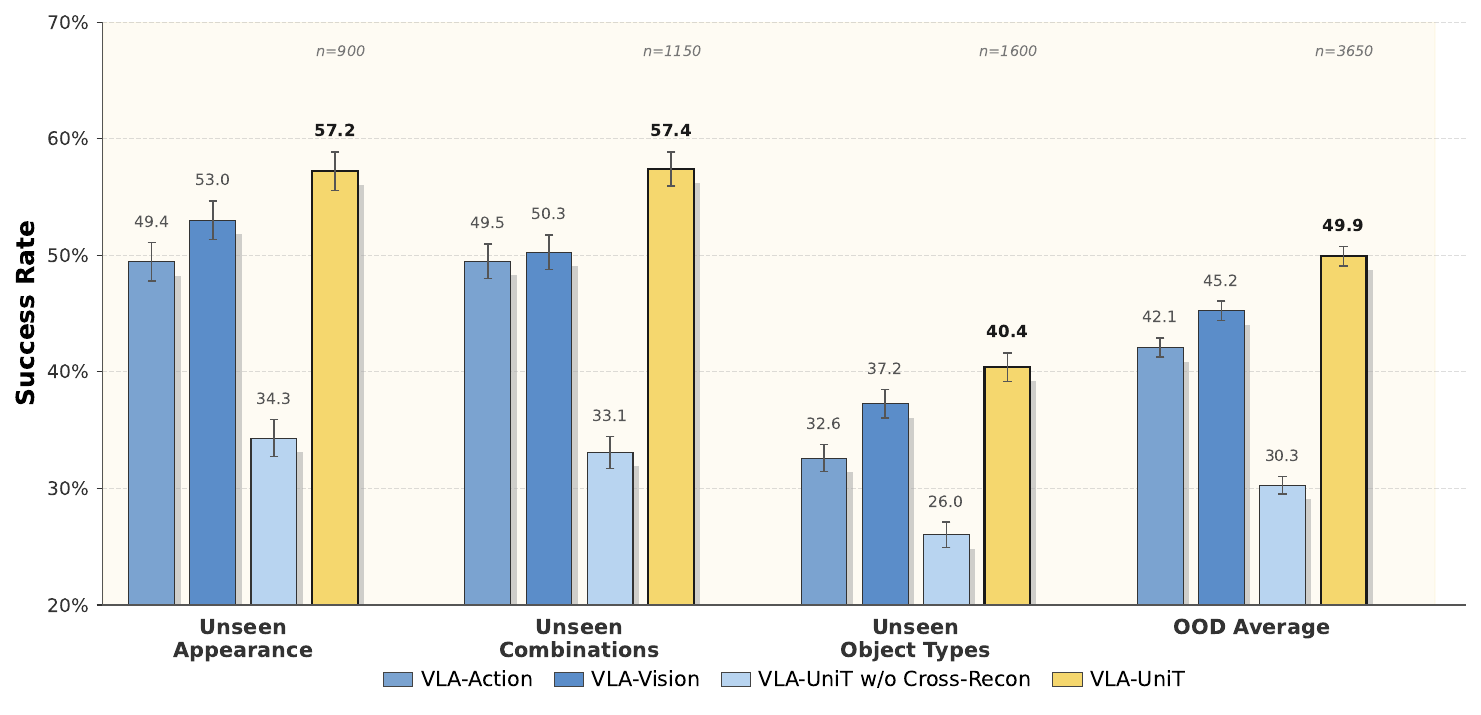}
    \hfill
    \includegraphics[width=0.36\textwidth]{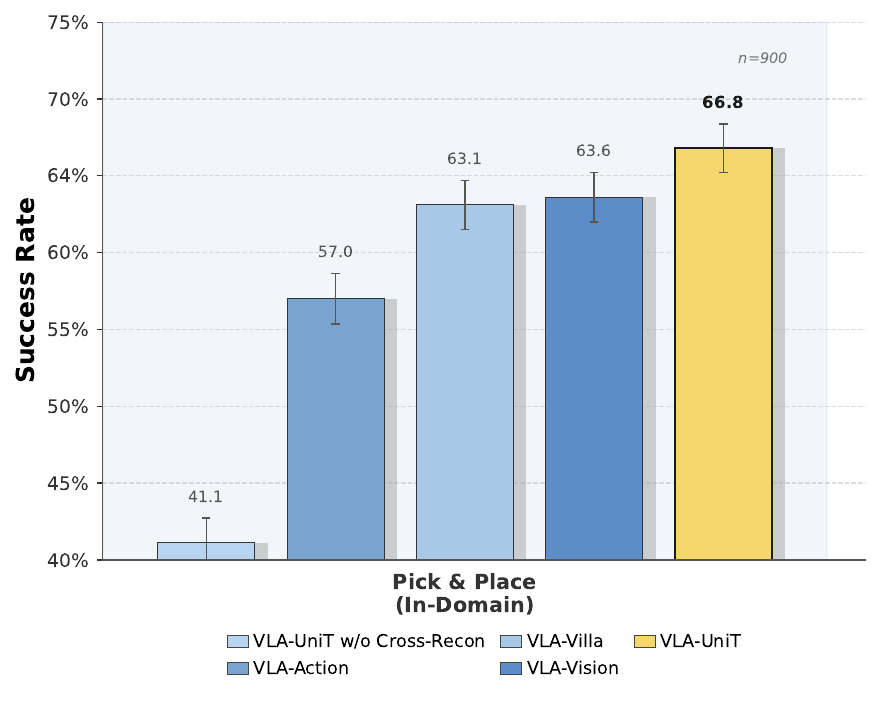}
    \caption{
    \textbf{Tokenizer paradigm ablation} on RoboCasa GR1 (full data with human co-training). \textbf{(Left)} OOD generalization. \textbf{(Right)} In-domain Pick \& Place performance.
    }
    \label{fig:ablation}
\end{figure}

UniT's design rests on two key claims made in Sec.~\ref{sec:unit}: (1) both vision and action are needed — action-only methods suffer cross-embodiment distribution misalignment without visual grounding, while vision-only methods entangle low-level appearance and miss fine-grained motor detail; and (2) the two modalities must be explicitly aligned through cross-reconstruction, rather than treated as disconnected vocabularies. We validate both claims under the human-humanoid co-training setup (EgoDex + RoboCasa pre-train, RoboCasa fine-tune), where the tokenizer's ability to bridge embodiments is directly tested. We compare paradigms corresponding to the architectures in Fig.~\ref{fig:related_work}.

\paragraph{Vision-Action Synergy Enables Transfer.}
As shown in Fig.~\ref{fig:ablation} (left), VLA-UniT (OOD average 49.9\%) consistently outperforms both single-modality variants across all OOD scenarios. VLA-Vision (45.2\%) provides a transferable visual signal but lacks fine-grained motor detail; VLA-Action (42.1\%) captures motor intent but struggles with the cross-embodiment distribution gap without visual grounding. The joint encoding of both modalities in VLA-UniT combines the embodiment-invariant nature of vision with the precision of action, forming a more complete \textit{unified physical language} for cross-embodiment transfer.

\paragraph{Cross-Reconstruction Produces Aligned Representations.}
VLA-UniT w/o Cross-Recon (30.3\%) falls below even the single-modality variants despite having access to both modalities, showing that multi-modal input alone does not guarantee alignment. VLA-UniT's cross-reconstruction objective addresses this by enforcing mutual reconstruction between vision and action (Sec.~\ref{sec:unit}), transforming disconnected modalities into a coherent shared vocabulary. The resulting 19.6\% gain over the ablation without cross-reconstruction confirms that explicit cross-modal alignment is the key enabler for human-to-humanoid transfer.

\paragraph{Bidirectional vs.\ Unidirectional Reconstruction.}
On in-domain performance (Fig.~\ref{fig:ablation}, right), we further include VLA-Villa, which uses unidirectional V2A reconstruction. VLA-UniT (66.8\%) consistently outperforms VLA-Villa (63.1\%), confirming that bidirectional cross-reconstruction is more effective than unidirectional alternatives for producing aligned cross-embodiment tokens.

\section{Conclusion and Discussion}


We presented UniT, a visual-anchored latent action tokenizer that establishes a unified physical language for human-to-humanoid transfer through cross-reconstruction. In VLA-UniT, UniT improves policy performance and data efficiency, while enabling effective human-to-humanoid transfer with OOD generalization and zero-shot task transfer. In WM-UniT, UniT provides a stronger conditioning interface for cross-embodiment dynamics modeling and human-to-humanoid transfer. Ablation studies confirm that both vision-action synergy and bidirectional cross-reconstruction are essential for these gains.

Looking forward, the visual branch of UniT encodes physical transitions from observations alone, without requiring paired action annotations. This opens a path toward absorbing the vast and largely untapped reservoir of internet video, where humans perform diverse physical tasks without motor labels. Such data could serve as an additional source of physical priors that enriches the shared latent space. Furthermore, the fact that UniT serves as a unified interface for both policy and world model suggests a deeper possibility: policies can propose latent actions, world models can simulate their visual consequences, and the resulting imagined rollouts can flow back as reward signals for reinforcement learning or enable test-time planning through search over the latent space. This closed-loop co-evolution, mediated entirely within the shared token space, may be a compelling route toward scalable embodied intelligence. On the data side, UniT's data-driven alignment readily scales to broader internet-scale human motion data without manual kinematic correspondence, enabling the framework to fully leverage massive heterogeneous datasets. This scalability also holds the potential to learn upper-body coordination and dexterous control directly from diverse human demonstrations.

\section*{Acknowledgments}
\small
We thank Chuan Ma and Lu Qiu for sharing the codebases for world model experiments, and Hui Zhou for his help with the real robot infrastructure and teleoperation data.

\bibliography{neurips_2026}
\bibliographystyle{unsrt}

\end{document}